\documentclass{article}

\usepackage[utf8]{inputenc}
\usepackage{newunicodechar}
\newunicodechar{♫}{\textmusicalnote}

\PassOptionsToPackage{numbers, compress}{natbib}

\usepackage[preprint]{neurips_2025}

\usepackage[utf8]{inputenc} 
\usepackage[T1]{fontenc}    
\usepackage{hyperref}       
\usepackage{url}            
\usepackage{booktabs}       
\usepackage{amsfonts}       
\usepackage{nicefrac}       
\usepackage{microtype}      
\usepackage{xcolor}         
\usepackage{amsmath}
\usepackage{graphicx}
\usepackage{algorithm}
\usepackage{algpseudocode}
\usepackage{subcaption}
\usepackage{pifont}
\usepackage{xspace}
\usepackage{threeparttable}
\usepackage{tcolorbox}
\usepackage{caption}

\tcbset{
  mypromptbox/.style={
    colback=gray!10,
    colframe=black,
    boxrule=0.8pt,
    arc=2pt,
    left=6pt,
    right=6pt,
    top=6pt,
    bottom=6pt,
    fonttitle=\bfseries,
    fontupper=\small,
    coltitle=white,
    colbacktitle=black,
  }
}

\newcommand{\OURS}{ComposeRAG\xspace}
\newcommand{\tref}[1]{Table.~\ref{#1}}
\newcommand{\fref}[1]{Fig.~\ref{#1}}
\newcommand{\aref}[1]{Algorithm.~\ref{#1}}
\newcommand{\apref}[1]{Appendix~\ref{#1}}

\newcommand{\sref}[1]{Section~\ref{#1}}

\newcommand{\cmark}{\ding{51}} 
\newcommand{\xmark}{\ding{55}} 
\newcommand{\ocircle}{\ding{109}}  

\title{\OURS: A Modular and Composable RAG for Corpus-Grounded Multi-Hop Question Answering}

\author{
  Ruofan Wu\thanks{Work done during an internship with Snowflake AI Research.}\\ 
  University of Houston\\
  \texttt{rwu13@cougarnet.uh.edu} \\
  \And
  Youngwon Lee \\
  Seoul National University \\
  \texttt{ywlee@ldi.snu.ac.kr} \\
  \And
  Fan Shu \\
  University of Houston \\
  \texttt{fshu@cougarnet.uh.edu} \\
  \And
  Danmei Xu \\
  Snowflake AI Research \\
  \texttt{danmei.xu@snowflake.com} \\
  \And
  Seung-won Hwang \\
  Seoul National University \\
  \texttt{seungwonh@snu.ac.kr} \\
  \And
  Zhewei Yao\thanks{Project Lead. The code is released at \href{https://github.com/Snowflake-Labs/Arctic_Agentic_RAG}{Arctic Agentic RAG}.} \\
  Snowflake AI Research \\
  \texttt{zhewei.yao@snowflake.com} \\
  \And
  Yuxiong He \\
  Snowflake AI Research \\
  \texttt{yuxiong.he@snowflake.com} \\
  \And
  Feng Yan \\
  University of Houston \\
  \texttt{fyan5@central.uh.edu} \\
}

\begin{document}

\maketitle

\begin{abstract}
Retrieval-Augmented Generation (RAG) systems are increasingly diverse, yet many suffer from monolithic designs that tightly couple core functions like query reformulation, retrieval, reasoning, and verification. This limits their interpretability, systematic evaluation, and targeted improvement, especially for complex multi-hop question answering. We introduce \textbf{\OURS}, a novel modular abstraction that decomposes RAG pipelines into atomic, composable modules. Each module, such as Question Decomposition, Query Rewriting, Retrieval Decision, and Answer Verification, acts as a parameterized transformation on structured inputs/outputs, allowing independent implementation, upgrade, and analysis. To enhance robustness against errors in multi-step reasoning, \OURS incorporates a self-reflection mechanism that iteratively revisits and refines earlier steps upon verification failure. Evaluated on four challenging multi-hop QA benchmarks, \OURS consistently outperforms strong baselines in both accuracy and grounding fidelity. Specifically, it achieves up to a $15\%$ accuracy improvement over fine-tuning-based methods and up to a $5\%$ gain over reasoning-specialized pipelines under identical retrieval conditions. Crucially, \OURS significantly enhances grounding: its verification-first design reduces ungrounded answers by over $10\%$ in low-quality retrieval settings, and by approximately $3\%$ even with strong corpora. Comprehensive ablation studies validate the modular architecture, demonstrating distinct and additive contributions from each component. These findings underscore \OURS's capacity to deliver flexible, transparent, scalable, and high-performing multi-hop reasoning with improved grounding and interpretability.
\end{abstract}
\section{Introduction}
Large Language Models (LLMs), including prominent examples like GPT-3, GPT-4, and LLaMA \cite{brown2020language, achiam2023gpt, touvron2023llama}, have demonstrated remarkable proficiency in various NLP tasks. Nevertheless, their reliance on static, pre-trained knowledge renders them susceptible to generating inaccuracies (hallucinations) \cite{ji2023survey} and limits their access to up-to-date or domain-specific information \cite{guu2020retrieval}. Retrieval-Augmented Generation (RAG) \cite{lewis2020retrieval} has emerged as a critical paradigm to overcome these deficiencies by integrating external knowledge sources, thereby enhancing factual grounding. While foundational RAG approaches (e.g., RETRO \cite{borgeaud2022improving}, RE-RAG \cite{kim2024re}) have advanced single-step QA, they encounter significant challenges when confronted with complex multi-hop queries. Such queries necessitate not only information retrieval but also sophisticated processes of decomposition and step-by-step reasoning across multiple documents—a capability that remains a substantial hurdle for conventional RAG frameworks.

The intricacies of multi-hop QA have spurred the development of specialized pipeline strategies. Initial attempts, including chain-of-thought prompting \cite{wei2022chain} and dedicated systems like MDR \cite{xiong2020answering} and TTQA-RS \cite{bardhan2024ttqa}, aimed to structure this complex reasoning. However, these early solutions often exhibited rigidity and were susceptible to error propagation from initial steps. While subsequent systems such as RQ-RAG \cite{chan2024rq} and Search-o1 \cite{li2025search} introduced greater flexibility, they also presented new challenges, including dependencies on supervised fine-tuning or inadequate verification mechanisms that result in poorly supported answers. A common deficiency across many existing multi-hop QA systems is their tendency towards monolithic or opaque architectures, which fundamentally limits their interpretability, adaptability, and potential for systematic enhancement.

In this work, we propose a modular and composable multi-hop RAG pipeline, \textbf{\OURS}, that emphasizes grounding, transparency, and flexibility. Our framework consists of independently interchangeable modules—such as question decomposition, passage reranking, query rewriting, retrieval decision, and answer verification—coordinated through prompt-based interfaces. To enhance robustness in complex scenarios, we incorporate a self-reflection mechanism that enables the system to iteratively revise its reasoning steps based on past outputs. Unlike prior systems, our approach explicitly enumerates and modularizes each component of the multi-hop reasoning process. This enables fine-grained analysis, targeted upgrades, and flexible adaptation to new tasks or domains. 

We evaluate \OURS on four multi-hop QA benchmarks—HotpotQA, 2WikiMultiHopQA, MuSiQue, and Bamboogle—and compare against strong baselines including RQ-RAG and Search-o1 across multiple LLM backbones. Our framework consistently outperforms these baselines in both accuracy and grounding fidelity, achieving up to a $15\%$ improvement in overall accuracy compared to fine-tuning-based methods and a $5\%$ gain over reasoning-specialized pipelines under identical retrieval conditions. It generates more faithfully grounded answers, enables efficient early exits, and scales effectively across model sizes. Notably, \OURS’s verification-first design reduces ungrounded answers by over $10\%$ in low-quality retrieval settings, and by approximately $3\%$ even when high-quality corpora are used. Comprehensive ablation studies confirm the value of modularity, revealing distinct and additive contributions from each module. Together, these results demonstrate \OURS’s ability to deliver flexible, interpretable, and high-performing multi-hop reasoning with improved factual grounding.

\begin{figure}[htbp]
  \centering
  \includegraphics[width=\linewidth]{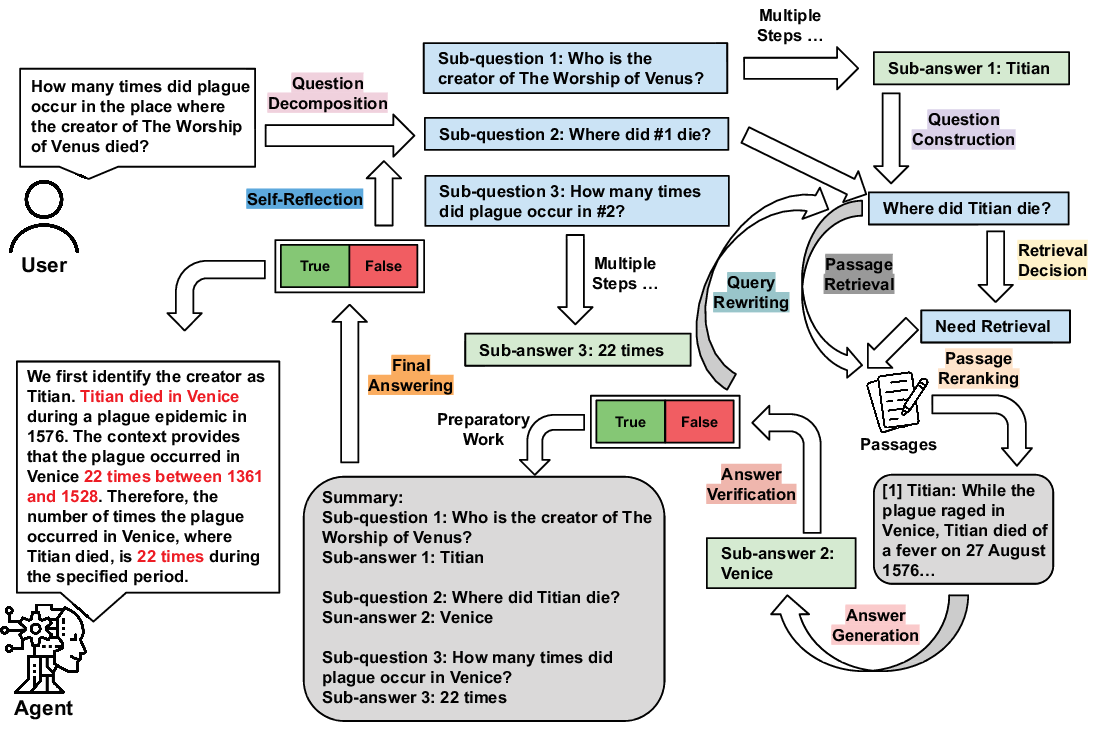}
  \caption{An overview of our modular Multi-Hop QA pipeline, highlighting the use of specialized modules to solve complex questions via step-by-step reasoning.}
  \label{main_process}
\end{figure}

Our key contributions include:

\begin{itemize}
\item \textbf{Modular Architecture:} A flexible and interpretable design for Multi-Hop QA that supports plug-and-play reasoning components, allowing for transparent analysis and targeted improvements (\sref{sec:composerag}).
\item \textbf{Orchestration with Self-Reflection:} An iterative orchestration strategy from self-reflecting  question decomposition and other reasoning steps to recover from early errors and enhance overall robustness, as validated by performance gains with increased reflection steps (\sref{ablation: self}).
\item \textbf{Well-composed:} With these contributions, we report empirical   studies that confirm ComposeRAG is well-composed by:
\begin{itemize}
\item Quantifying the distinct and additive performance contributions of core modules such as Question Decomposition, Passage Reranking, and Answer Verification (\sref{ablation: add}).
\item Confirming independent module upgradability, where enhancing individual components (e.g., by leveraging more powerful LLMs like GPT-4o over GPT-4o-mini for specific tasks) in isolation yields measurable improvements, showcasing the system's scalability and adaptability (\sref{ablation: upgrade}).
\item Underscoring the positive impact of efficiency-oriented components like the Retrieval Decision module and the Simple QA pipeline on performance and resource utilization (\sref{ablation: efficiency}).
\end{itemize}
\end{itemize}
\section{Related Work}

The development of \OURS builds upon several key areas of research in question answering and information retrieval.

\textbf{RAG for Open-Domain QA:}
LLMs for open-domain QA
often leads to issues with factual consistency and the recall of specific or infrequently encountered information. 
RAG \cite{lewis2020retrieval} was introduced to mitigate these limitations by enabling LLMs to access and condition their generation on information retrieved from external corpora. For instance, Atlas \cite{izacard2023atlas} demonstrated improved few-shot learning by combining a dense retriever with a sequence-to-sequence generator. Despite these advancements, RAG systems employing a single retrieval pass often struggle with complex questions that require synthesizing information from multiple documents through step-by-step reasoning.

\textbf{Multi-Hop QA:}
Addressing the demands of multi-hop questions, which require reasoning across multiple pieces of evidence, has led to specialized techniques focusing on structured reasoning and query decomposition. Methods like GenDec \cite{wu2024gendec} generate independent, self-contained sub-questions that can be answered individually, thereby reducing the risk of error propagation inherent in sequential multi-step reasoning. Concurrently, significant effort has been directed towards enhancing the quality of retrieved information. RankGPT \cite{sun2023chatgpt}, for example, utilizes the zero-shot capabilities of LLMs to re-rank retrieved passages based on instructional prompts, prioritizing more relevant documents. Similarly, SuRe \cite{kim2024sure} improves pipeline robustness by summarizing retrieved passages for each candidate answer, assessing the level of support, and ultimately selecting the most well-grounded response. These approaches highlight the dual importance of effective decomposition and high-quality evidence in tackling multi-hop QA.

\textbf{RAG Reasoning Methods:}
Closest line of work to ours is 
recent RAG systems,  tightly integrating reasoning capabilities. RQ-RAG \cite{chan2024rq} fine-tunes a query rewriting module to improve multi-hop retrieval quality, enabling more accurate downstream generation but requiring task-specific training. In contrast, Search-o1 \cite{li2025search} adopts an agentic approach, prompting large reasoning models to iteratively query search engines and synthesize answers. ReARTeR \cite{sun2025rearter} introduces a trust-based reward model that reinforces trustworthy intermediate reasoning steps. R1-Searcher \cite{song2025r1} applies reinforcement learning in two stages to train LLMs to decide when and how to search effectively. Search-R1 \cite{jin2025search} further integrates multi-turn search and reasoning through reinforcement learning with outcome-based rewards and retrieved token masking.

\textbf{Our Distinction:}
We propose a composable RAG reasoning framework, ComposeRAG,
where each module positively contributes to performance and can be independently upgraded for further gains. Modules are orchestrated by
\textbf{self-reflection}, to
 iteratively reassess, refine, and optimize  retrieval and reasoning processes. We will further distinguish our framework in Section 3.2.
 
 A notable example of self-reflection is self-ask prompting \cite{press2022measuring}, where LLMs are guided to generate and answer intermediate sub-questions before arriving at a final answer, fostering more coherent and transparent reasoning. EfficientRAG \cite{zhuang2024efficientrag} builds on this by introducing an iterative retrieval strategy that generates follow-up queries without repeated LLM invocations, thus minimizing computational costs while filtering irrelevant content. AUTO-RAG \cite{yu2024auto} extends this concept by framing retrieval as a multi-turn dialogue, enabling the model to adaptively refine its queries and determine when sufficient information has been gathered. 
 Unlike methods that focus on query refinement or sub-question planning, we aim to integrate self-reflection with the full reasoning pipeline.
\section{Motivations}
\label{motivation}

While recent systems such as RQ-RAG \cite{chan2024rq} and Search-o1 \cite{li2025search} have advanced the state of multi-hop question answering (QA), they contend with critical limitations that hinder their broader applicability and robustness:

First, \textbf{Rigidity in Component Management and Upgradability.} Existing systems, such as RQ-RAG, often employ monolithic designs where core modules are tightly coupled. This structural rigidity makes it difficult to independently substitute, upgrade, or analyze individual components. Consequently, experimental agility is reduced, and adapting the pipeline to new tasks or leveraging improved individual models becomes a substantial re-engineering effort. This underscores the need for a truly modular architecture that supports independent component enhancement, a key aspect of \OURS's design and validated by our ablation studies (\sref{ablation: upgrade}).

Second, \textbf{Deficits in Modularity and Reasoning Transparency.} While some systems like Search-o1 demonstrate strong multi-step reasoning capabilities, their architectures may not explicitly isolate or formalize the distinct stages of the reasoning process. This lack of explicit modularity can obscure internal decision-making, making it challenging to pinpoint sources of error, attribute them to specific functional blocks, or systematically diagnose reasoning flaws. As a result, system interpretability is diminished, especially when analyzing failures. This highlights the importance of a transparent, modular framework like \OURS, where each step is an identifiable and analyzable component (see \sref{sec:core_modules} and \sref{sec:pipelines} for design and \sref{ablation: add} for validation).

Third, \textbf{Insufficient Grounding and Lack of Robust Verification Loops.} Advanced reasoning pipelines can sometimes produce plausible-sounding answers that lack adequate grounding in the retrieved evidence, as observed in systems like Search-o1. The absence of rigorous verification mechanisms at each step, or the inability to iteratively refine reasoning based on verification failures, elevates the risk of factual hallucinations and undermines trustworthiness. This points to the necessity for systems that not only verify answers but can also engage in self-correction—a core feature of \OURS through its Answer Verification module and Self-Reflection Mechanism (see \sref{sec:core_modules} and \sref{sec:self_reflection_mechanism} for design and \sref{ablation: add} for validation).

These limitations collectively highlight the need for a Multi-Hop QA framework that is inherently modular, transparent, and robust. Such a framework should empower researchers and practitioners with the flexibility to manage and upgrade components independently, offer clear insights into the reasoning process for effective debugging and improvement, ensure strong factual grounding through systematic verification, and incorporate mechanisms for iterative self-correction. \OURS is designed to embody these principles.

\section{\OURS}
\label{sec:composerag}

Addressing the limitations of monolithic and opaque RAG systems outlined in \sref{motivation}, we introduce \textbf{\OURS}. This framework embodies a modular and composable approach tailored for the complexities of multi-hop question answering. The foundational principle of \OURS is the decomposition of the intricate multi-hop reasoning process into a series of atomic, interchangeable modules. Each module is conceptualized as a parameterized transformation, operating on structured inputs to produce structured outputs. This design facilitates independent implementation, rigorous analysis, and targeted upgrades of individual components.

To ensure meaningful modularity, we adopt the formal notion of a \emph{well-composed system}, where each module $M_i$ contributes measurably to system performance. Specifically, we assess:
\begin{itemize}
  \item \textbf{Ablation:} Removing $M_i$ yields a drop in performance $P$:
  $\Delta P_{-i} = P(S) - P(S \setminus M_i) \ge 0$.
  \item \textbf{Upgrade:} Substituting $M_i$ with a better version $M_i'$ improves performance: 
  $\Delta P_{+i} = P(S[M_i \leftarrow M_i']) - P(S) > 0$.
\end{itemize}
These two properties, denoted as \textbf{P1} and \textbf{P2} guide our study in \sref{ablation}, validating 
both properties hold in ComposeRAG, in \sref{ablation: add} and \sref{ablation: upgrade}, respectively:
Well-composed system offers granular control by isolating the functionality of each module, and allows for precise interventions when errors occur.

\subsection{Composable Building Blocks: The Core Modules of \OURS}

\label{sec:core_modules}

The \OURS framework is constructed from a suite of specialized, yet composable, modules, each engineered to perform a distinct function within the overarching question answering process. These modules are systematically categorized based on their primary contribution to the reasoning pipeline, encompassing decomposition, evidence handling, and verified answering.

\subsubsection{Decomposition and Stepwise Reasoning}
\begin{figure}[htbp]
  \centering
  \includegraphics[width=\linewidth]{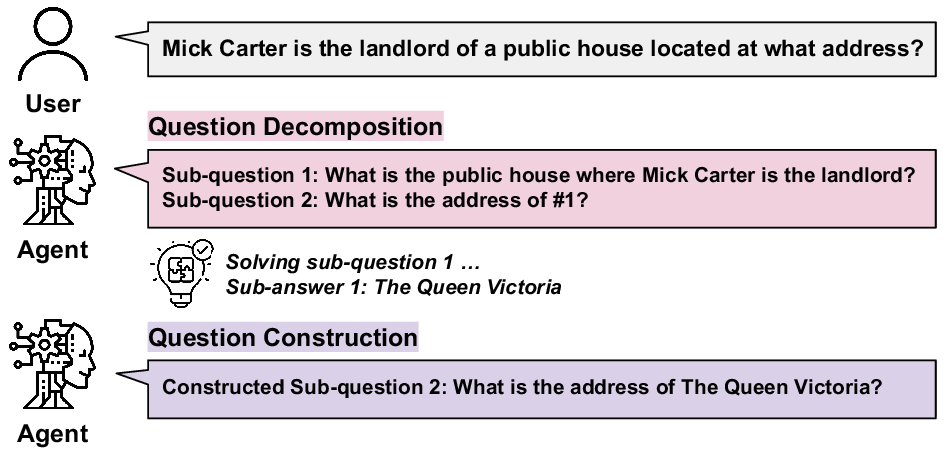}
  \caption{Question Decomposition and Construction Example}
  \label{fig:qd_qc_example}
\end{figure}

Central to tackling multi-hop questions is the ability to deconstruct a complex question into a sequence of manageable sub-questions. The \textbf{Question Decomposition} module is responsible for this initial transformation, linearly breaking down the input question into simpler, interrelated sub-questions. Unlike approaches that enforce rigid tree-like structures, \OURS employs a flexible placeholder-based annotation system to signify dependencies between these sub-questions. Subsequently, the \textbf{Question Construction} module takes these decomposed, placeholder-annotated sub-questions and instantiates them into actionable queries by filling in the placeholders with answers derived from preceding reasoning steps. This ensures that each sub-question presented to the retrieval and answering components is self-contained and contextually complete, as illustrated in the example (\fref{fig:qd_qc_example}).

\subsubsection{Evidence Retrieval and Prioritization}
\begin{figure}[htbp]
  \centering
  \includegraphics[width=\linewidth]{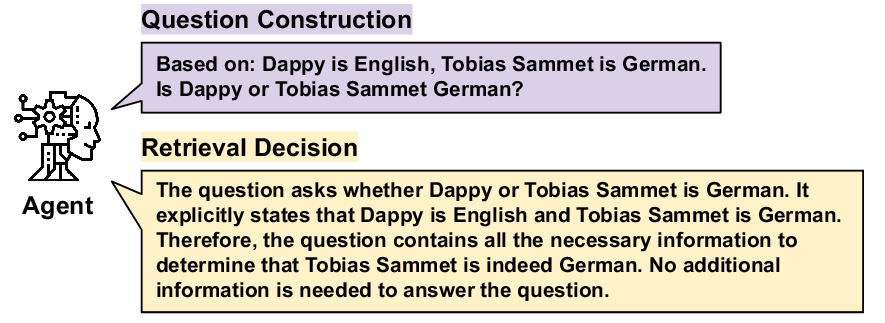}
  \caption{Retrieval Decision Example}
  \label{fig:rd_example}
\end{figure}

Effective reasoning is critically dependent on the retrieval and judicious prioritization of high-quality, relevant evidence. To this end, \OURS incorporates several modules. The \textbf{Retrieval Decision} module acts as an efficiency-enhancing gatekeeper, assessing whether the current (sub-)question already contains sufficient information for an answer or if external document retrieval is indeed necessary (see \fref{fig:rd_example} for an illustration).

When retrieval is deemed necessary, the \textbf{Query Rewriting} module can be invoked to enhance retrieval precision, particularly for under-specified or context-dependent sub-questions. It reformulates such queries into more complete and effective search terms, leveraging information from prior answers and the broader conversational context. Once a set of potentially relevant passages is retrieved, the \textbf{Passage Reranking} module, inspired by recent work such as \cite{sun2023chatgpt}, reorders these passages based on their specific relevance to the current sub-question. This is achieved using prompt-based scoring mechanisms to elevate the most pertinent information for subsequent processing.

\subsubsection{Answering with Evidence-Based Verification}
Ultimately, the goal is to generate faithful, well-grounded answers. The \textbf{Answer Generation} module synthesizes a candidate answer for the current sub-question, utilizing the prioritized retrieved passages and, in multi-hop scenarios, the accumulated history of prior sub-questions and their answers. A core tenet is to produce answers that are explicitly supported by the provided evidence. To ensure this, the \textbf{Answer Verification} module critically evaluates the generated answer for factual consistency and adequate grounding against the cited passages. If the supporting evidence is deemed insufficient, or if the answer is found to be ungrounded, this module can trigger corrective actions within the pipeline, such as re-attempting a previous step, or signal an abstention if a high-confidence, verifiable answer cannot be produced. Finally, for multi-hop questions, once all constituent sub-questions have been successfully processed and their answers verified, the \textbf{Final Answering} module aggregates these intermediate results to synthesize a comprehensive and coherent final answer to the original complex query. Formal definitions and detailed operational parameters for each of these modules are provided in \apref{appendix:modules}.

\subsection{Orchestrating Reasoning: The \OURS Pipelines}
\label{sec:pipelines}
\begin{figure}[htbp]
  \centering
  \includegraphics[width=\linewidth]{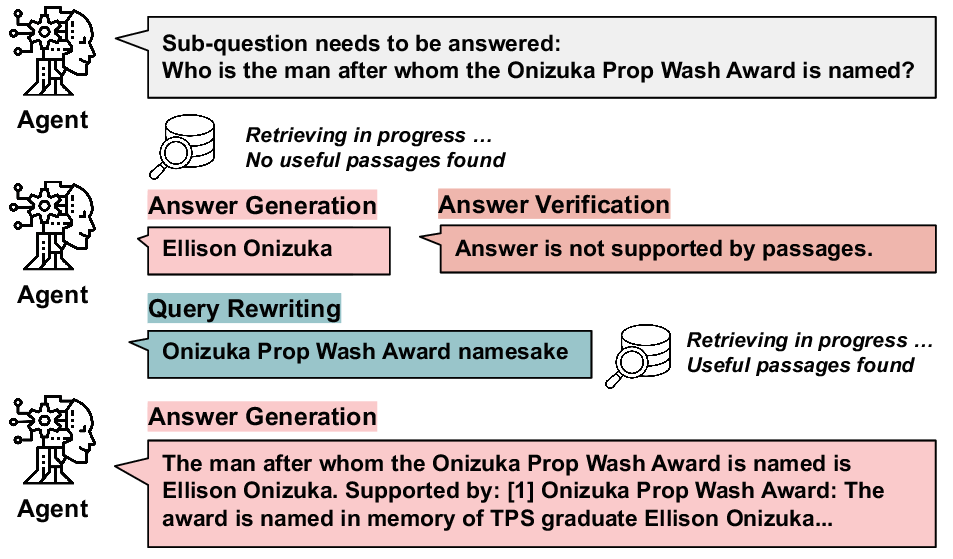}
  \caption{Iterative Refinement Example: Answer Verification and Query Rewriting}
  \label{fig:iterative_refinement_example}
\end{figure}

The core modules, as defined in \sref{sec:core_modules}, serve as the fundamental building blocks for two distinct, yet interconnected, question answering pipelines within the \OURS framework: the Simple QA Pipeline and the Advanced Multi-Hop QA Pipeline.

The \textbf{Simple QA Pipeline} is designed for straightforward questions that can typically be resolved through a single retrieval and generation cycle. This pipeline first retrieves relevant passages, then employs the Answer Generation module to synthesize an answer, and finally utilizes the Answer Verification module to validate its correctness and grounding. If the answer is successfully verified, it is returned to the user. If verification fails, or if the question is initially assessed as too complex for this direct approach, the system may escalate the query to the more comprehensive Multi-Hop QA Pipeline. Further implementation details of this streamlined pipeline are available in \apref{appendix:simple}.

For complex questions that necessitate multi-step reasoning and the integration of information from multiple sources, \OURS employs the \textbf{Advanced Multi-Hop QA Pipeline}, the general flow of which is illustrated in \fref{main_process}. This pipeline begins with the \textbf{Initial Decomposition} of the original complex question into a sequence of simpler sub-questions by the Question Decomposition module. Each of these sub-questions then undergoes an \textbf{Iterative Sub-Question Processing} cycle. Within this cycle, the Question Construction module first formulates an actionable query from the current sub-question and any previously derived answers. The Retrieval Decision module then determines if external evidence is required. If so, passages are retrieved—potentially enhanced by the Query Rewriting module for ambiguous or context-dependent sub-questions—and subsequently prioritized by the Passage Reranking module. Following this, the Answer Generation module produces a candidate sub-answer based on the provided evidence. Crucially, this sub-answer is then scrutinized by the Answer Verification module. If verification fails (e.g., due to insufficient evidence or a poorly formulated sub-query leading to irrelevant retrieval), this can trigger corrective actions, as demonstrated in the iterative refinement example (\fref{fig:iterative_refinement_example}).

Once all sub-questions have been successfully addressed and their answers verified through this iterative process, the \textbf{Final Synthesis} step occurs, where the Final Answering module aggregates the intermediate results into a single, comprehensive answer to the original multi-hop query. This structured, iterative, and verification-centric process is designed to enhance both the transparency and the robustness of the multi-hop reasoning.

\subsection{Orchestration: Self-Reflection}

In composable design, orchestration plays a critical role, and orchestration should be \textbf{progress-informed}, to adapt based on module outputs, or partial evaluation signals, for updating orchestration policies towards improving downstream performance.

For example,
if the initial breakdown of a complex question is flawed, these errors may propagate through subsequent steps, potentially leading to an incorrect or ungrounded final answer. Desirably, orchestration should be informed and proactively mitigate this, \OURS incorporates a \textbf{Self-Reflection Mechanism}. This mechanism allows the entire pipeline to revisit and refine its reasoning process if the final synthesized answer fails a concluding verification check (the overall process is summarized in \aref{self-reflection}).

The self-reflection process operates through several stages. First, the final answer produced by the Multi-Hop QA Pipeline undergoes a rigorous check by the Answer Verification module. If this final verification fails, the system initiates an \textbf{Error Analysis and Diagnosis} phase. During this phase, it analyzes the complete reasoning trace, which includes the initial decomposition, all intermediate sub-questions, the retrieved passages, and the generated sub-answers, to identify the most probable source of the error. Often, such errors can be traced back to a suboptimal initial decomposition. Based on this diagnosis, specialized prompting techniques are used to generate \textbf{Guided Re-decomposition} instructions. These instructions provide corrective feedback to the Question Decomposition module, guiding it on how to improve the initial breakdown of the original question. The Question Decomposition module then re-processes the original question, now informed by this targeted feedback, to produce a revised set of sub-questions. Finally, the Multi-Hop QA Pipeline is \textbf{Re-executed} using this new, hopefully improved, set of sub-questions.

This capacity for iterative refinement at the pipeline level allows \OURS to learn from its mistakes within the context of a single problem-solving instance. By diagnosing and correcting flaws in its own reasoning process, particularly in the critical decomposition phase, the self-reflection mechanism significantly improves the reliability and accuracy of the final answers, demonstrating the adaptive power inherent in \OURS's modular design.

\label{sec:self_reflection_mechanism}
\begin{algorithm}[H]
\caption{Self-Reflection for Decomposition Improvement}
\label{self-reflection}
\begin{algorithmic}[0]
\State \textbf{Specification:} Final answer \( a_{\text{final}} \), question \( q \), supporting passages \( \mathcal{P} \), question solving record \( H \)
\State \( (y, r) \gets \mathrm{Verify}_\theta(q, a_{\text{final}}, \mathcal{P}) \) \Comment{Step 1: Final answer verification}

\If{ \( y = \texttt{True} \) }
    \State \Return \( a_{\text{final}} \) \Comment{Answer is correct}
\Else
    \State \( \text{analysis} \gets \mathrm{ImproveAnalysis}_\theta(q, H) \) \Comment{Step 2: Analyze the solving record}

    \State \( \mathbf{S}_{\text{new}} \gets \mathrm{ImproveDecomposition}_\theta(q, \text{analysis}) \) \Comment{Step 3: Generate new decomposition}

    \State \( a_{\text{new}} \gets \mathrm{MultiHop QA pipeline}(q, \mathbf{S}_{\text{new}}) \) \Comment{Step 4: Rerun the pipeline}

    \State \Return \( a_{\text{new}} \)
\EndIf
\end{algorithmic}
\end{algorithm}

\section{Experiments}
\subsection{Experimental Setup}
\label{sec:experimental_setup}

\textbf{Datasets and Retrieval Corpus:} We evaluate \OURS on four widely-used multi-hop question answering datasets: HotpotQA \cite{yang2018hotpotqa}, 2WikiMultiHopQA \cite{ho2020constructing}, MuSiQue \cite{trivedi2022musique}, and Bamboogle \cite{press2022measuring}. These datasets are designed to test complex reasoning capabilities by requiring models to integrate information from multiple sources. For HotpotQA, 2WikiMultiHopQA, and MuSiQue, we randomly sampled 500 questions from their respective dev sets for evaluation. For Bamboogle, we used the entire test set comprising 125 questions.
For open-domain retrieval, we utilize the English Wikipedia corpus as the background knowledge source. We employ two versions of this corpus:
\begin{itemize}
    \item A truncated version of the 2018-12-20 Wikipedia dump, as released by FlashRAG \cite{jin2024flashrag}, where articles are segmented into short chunks.
    \item A version pre-processed by us using the 2019-08-01 Wikipedia dump from the KILT knowledge source \cite{petroni2020kilt}. In our pre-processing, we retain the original abstract of each article and segment the main body into chunks of 4-5 sentences to ensure consistency and relevance for dense retrieval.
\end{itemize}

\textbf{Baselines:} 

We aim to evaluate multi-hop QA systems within a modular reasoning framework that emphasizes transparency, component-level control, and iterative verification. To this end, we focus on systems that align with these design principles and are compatible with our evaluation protocol. We select \textbf{RQ-RAG} and \textbf{Search-o1} as our primary baselines for reproduction and comparison. These two methods represent complementary paradigms—RQ-RAG employs supervised fine-tuning for query refinement, while Search-o1 leverages zero-shot agentic prompting with large language models.
\begin{itemize}
    \item \textbf{RQ-RAG \cite{chan2024rq}:} We follow the setup in FlashRAG \cite{jin2024flashrag}, using e5-base-v2 as the embedding model, to reproduce its results.
    \item \textbf{Search-o1 \cite{li2025search}:} It originally relies on Bing Search combined with Jina AI for real-time web retrieval. To ensure a fair comparison with our system focused on a local Wikipedia corpus, we adapt its retrieval mechanism to use the same local corpus as \OURS.
\end{itemize}

Although we do not re-implement reinforcement learning-based methods, we include \textbf{ReARTeR} \cite{sun2025rearter} and \textbf{R1-Searcher} \cite{song2025r1} by reporting their published results, as they share similar evaluation setups and address related reasoning objectives. \textbf{Search-R1} \cite{jin2025search}, on the other hand, focuses on learning reasoning and retrieval policies via reinforcement learning with relatively small-scale models, and adopts a tightly coupled end-to-end architecture. As its design goals and evaluation protocol differ substantially from ours, we do not include it in the main set of empirical comparisons, but discuss it in related work for completeness.

\textbf{Implementation Settings:} For LLM-based generation within \OURS, we primarily use three series of models:
\begin{itemize}
    \item GPT-based models (e.g., GPT-4o, GPT-4o-mini) accessed via the Azure OpenAI Service, providing reliable and scalable inference. (Note: Citations for GPT-4o and mini might be more recent than \cite{achiam2023gpt} for GPT-4, adjust if necessary).
    \item Models based on LLaMA 3.1 \cite{grattafiori2024llama}, accessed using the Cortex Completion Service for efficient inference with open-weight models.
    \item Locally deployed models such as QwQ-32B \cite{zheng2024processbench} and Qwen2.5 \cite{yang2024qwen2}, using the vLLM framework on 8xV100 GPUs.
\end{itemize}
For dense retrieval, \OURS supports two configurations:
\begin{itemize}
    \item The e5-base-v2 model \cite{wang2022text} to encode queries and passages into dense vector representations, with indexing handled by FAISS \cite{douze2024faiss} for efficient similarity-based retrieval.
    \item The Cortex Search API, which utilizes snowflake-arctic-embed-m-v1.5 as its default embedding model.
\end{itemize}

\textbf{Evaluation Metrics:} We employ two primary metrics to evaluate performance:
\begin{itemize}
    \item \textbf{LLM Evaluation:} We use GPT-4o as an automatic evaluator to judge whether a predicted answer semantically aligns with the ground truth. The evaluation is guided by a structured prompt that considers relevance, semantic consistency, and completeness. The model outputs true if the answer matches any ground truth in meaning and contains all key information, and false otherwise. This setup aligns with recent practices in using LLMs as evaluators \cite{gu2024survey}.
    \item \textbf{Cover Exact Match (Cover EM):} This metric is a relaxed variant of the standard Exact Match. Let $\hat{a}$ be the predicted answer and $\mathcal{A}=\{a_{1},a_{2},...,a_{k}\}$ be the set of ground truth answers. For each answer string, we apply a normalization function, $\text{Normalize}()$, which lowercases the text, removes punctuation, trims whitespace, and tokenizes the result. Let $T_{\hat{a}}=\text{Normalize}(\hat{a})$ be the tokenized prediction, and $T_{a_{j}}=\text{Normalize}(a_{j})$ be the tokenized form of each ground truth $a_{j}\in\mathcal{A}$. The function $\text{Subseq}(s,t)$ returns 1 if $s$ is a contiguous subsequence of $t$, and 0 otherwise. Cover EM is then computed as:
    $$\mathrm{CoverEM}(\hat{a}, \mathcal{A}) =
    \begin{cases}
    1, & \text{if } \exists a_j \in \mathcal{A} \text{ such that } \mathrm{Subseq}(T_{a_j}, T_{\hat{a}}) = 1 \\
    0, & \text{otherwise}
    \end{cases}$$
    This metric accounts for predictions that contain longer or more natural language expressions, as long as the essential answer span from a ground truth appears intact within the prediction.
\end{itemize}

\subsection{Empirical Results and Analysis}
\label{sec:empirical_results}

This section presents the empirical evaluation of \OURS, focusing on its performance against established baselines and the scalability of its modular design. The primary results across four multi-hop QA datasets are summarized in \tref{results-table}. A detailed discussion regarding model selection, the compatibility of models with our pipeline's design, and the strategy for ensuring fair comparisons—including the alignment of retrieval corpora and LLM pairings—is provided in \apref{appendix:comparison}.

\begin{table}[htbp]
  \caption{Results table for four multi-hop datasets with different retrieval and generation setting}
  \label{results-table}
  \centering
  \small
  \begin{tabular}{lllllll}
    \toprule
    \multicolumn{7}{l}{\textbf{HotpotQA}} \\
    \midrule
    Method     & LLM     & Retriever     & Corpus     & LLM Eval     & Cover-EM    & Avg  \\
    \midrule
    RQ-RAG & fine-tuned llama2-7b & e5-base-v2 & wiki2018 & 0.326 & 0.258 & 0.292 \\
    Ours & Llama3.1-8b & Cortex Search & wiki2018 & \textbf{0.494} & \textbf{0.388} & \textbf{0.441} \\
    \midrule
    Search-o1 & QwQ-32B & Cortex Search & wiki2019 & 0.664 & 0.492 & 0.578 \\
    Ours & Qwen2.5-72B-Instruct & Cortex Search & wiki2019 & \textbf{0.714} & \textbf{0.554} & \textbf{0.634} \\
    \midrule
    Ours & GPT-4o-mini & Cortex Search & wiki2019 & 0.702 & 0.558 & 0.630 \\
    Ours & GPT-4o & Cortex Search & wiki2019 & \textbf{0.732} & \textbf{0.580} & \textbf{0.656} \\
    \midrule
    \multicolumn{7}{l}{\textbf{2WikiMultiHopQA}} \\
    \midrule
    Method     & LLM     & Retriever     & Corpus     & LLM Eval     & Cover-EM   & Avg  \\
    \midrule
    RQ-RAG & fine-tuned llama2-7b & e5-base-v2 & wiki2018 & 0.240 & 0.268 & 0.254 \\
    Ours & Llama3.1-8b & Cortex Search & wiki2018 & \textbf{0.326} & \textbf{0.334} & \textbf{0.330} \\
    \midrule
    Search-o1 & QwQ-32B & Cortex Search & wiki2019 & \textbf{0.808} & \textbf{0.770} & \textbf{0.789} \\
    Ours & Qwen2.5-72B-Instruct & Cortex Search & wiki2019 & 0.720 & 0.726 & 0.723 \\
    \midrule
    Ours & GPT-4o-mini & Cortex Search & wiki2019 & 0.740 & 0.728 & 0.734 \\
    Ours & GPT-4o & Cortex Search & wiki2019 & \textbf{0.820} & \textbf{0.808} & \textbf{0.814} \\ 
    \midrule
    \multicolumn{7}{l}{\textbf{MuSiQue}} \\
    \midrule
    Method     & LLM     & Retriever     & Corpus     & LLM Eval     & Cover-EM   & Avg  \\
    \midrule
    RQ-RAG & fine-tuned llama2-7b & e5-base-v2 & wiki2018 & 0.096 & 0.086 & 0.091 \\
    Ours & Llama3.1-8b & Cortex Search & wiki2018 & \textbf{0.128} & \textbf{0.102} & \textbf{0.115} \\
    \midrule
    Search-o1 & QwQ-32B & Cortex Search & wiki2019 & \textbf{0.374} & 0.278 & 0.326 \\
    Ours & Qwen2.5-72B-Instruct & Cortex Search & wiki2019 & 0.358 & \textbf{0.318} & \textbf{0.338} \\
    \midrule
    Ours & GPT-4o-mini & Cortex Search & wiki2019 & 0.376 & 0.328 & 0.352 \\
    Ours & GPT-4o & Cortex Search & wiki2019 & \textbf{0.388} & \textbf{0.342} & \textbf{0.365} \\
    \midrule
    \multicolumn{7}{l}{\textbf{Bamboogle}} \\
    \midrule
    Method     & LLM     & Retriever     & Corpus     & LLM Eval     & Cover-EM   &Avg  \\
    \midrule
    RQ-RAG & fine-tuned llama2-7b & e5-base-v2 & wiki2018 & 0.264 & 0.224 & 0.244 \\
    Ours & Llama3.1-8b & Cortex Search & wiki2018 & \textbf{0.432} & \textbf{0.384} & \textbf{0.408} \\
    \midrule
    Search-o1 & QwQ-32B & Cortex Search & wiki2019 & 0.696 & \textbf{0.648} & \textbf{0.672} \\
    Ours & Qwen2.5-72B-Instruct & Cortex Search & wiki2019 & \textbf{0.728} & 0.616 & \textbf{0.672} \\
    \midrule
    Ours & GPT-4o-mini & Cortex Search & wiki2019 & 0.712 & \textbf{0.640} & \textbf{0.676} \\
    Ours & GPT-4o & Cortex Search & wiki2019 & \textbf{0.728} & 0.624 & \textbf{0.676} \\
    \bottomrule
  \end{tabular}
\end{table}

\subsubsection{Comparison with RQ-RAG: General-Purpose Prompting vs. Task-Specific Fine-Tuning}
We first compare \OURS, when utilizing an instruction-tuned Llama3.1-8B model in a prompted configuration, against RQ-RAG, which employs a Llama2-7B model fine-tuned for the task. As evidenced in \tref{results-table}, \OURS demonstrates superior performance across all evaluated datasets without resorting to task-specific fine-tuning. Notably, on the HotpotQA and Bamboogle datasets, \OURS achieves substantial improvements in both LLM-based evaluation scores and Cover EM. For instance, on HotpotQA, \OURS (Llama3.1-8B) achieves an average score of $0.441$, compared to $0.292$ for RQ-RAG. This outcome highlights the efficacy of leveraging general-purpose, pre-trained instruction-tuned models within a well-structured modular prompting framework. Furthermore, these results underscore the inherent flexibility and scalability of the \OURS design, which circumvents the need for resource-intensive custom training pipelines for individual components or tasks, aligning with our goal of creating an adaptable and broadly applicable system.

\subsubsection{Comparison with Search-o1: Grounded Modular Reasoning vs. Ungrounded Specialized Reasoning}
Next, we benchmark \OURS, configured with the Qwen2.5-72B-Instruct model, against Search-o1, which utilizes the reasoning-specialized QwQ-32B model. To ensure an equitable comparison, both systems operate with the identical retriever (Cortex Search) and retrieval corpus (wiki2019). The results presented in \tref{results-table} indicate that \OURS achieves performance levels that are either superior to or comparable with Search-o1 across the majority of datasets. Specifically, on HotpotQA, \OURS (Qwen2.5-72B) shows an average score of 0.634 against Search-o1's $0.578$. Similarly, on Bamboogle, \OURS achieves an average of $0.672$, matching Search-o1's performance. On MuSiQue, \OURS demonstrates higher answer coverage (Cover-EM of $0.318$ vs. $0.278$), although its LLM evaluation score is marginally lower.

The primary exception to \OURS's strong performance is on 2WikiMultiHopQA, where Search-o1 outperforms it (average of $0.789$ vs. $0.723$ with Qwen2.5-72B). This dataset often involves implicit reasoning and sparse evidence, where Search-o1 tends to produce plausible but not always well-grounded answers. In contrast, \OURS emphasizes factual grounding through verification modules, which enhances correctness but can lower answer rates when sufficient supporting context is not retrieved. This trade-off is particularly relevant in applications requiring verifiable outputs.

To better understand this behavior, we analyze cases where Search-o1 succeeds but \OURS fails. We find that approximately $18\%$ of Search-o1-only correct answers lack direct support from the retrieved context. These answers are plausible but not explicitly grounded in evidence, raising concerns about their reliability. When averaged over the full dataset, these ungrounded yet correct responses account for $2.8\%$ of all evaluated examples. This analysis highlights a key tradeoff in our system: by prioritizing evidence-based verification, we may sacrifice coverage in favor of factual precision. Appendix~\ref{appendix:grounding} provides detailed examples and categorization of these grounding issues.

\subsubsection{Scalability Across Model Sizes: GPT-4o and GPT-4o-mini}
To evaluate how \OURS scales with the underlying model capacity, we test it using both GPT-4o and its lighter version, GPT-4o-mini. As shown in \tref{results-table}, increased model size consistently yields higher performance—GPT-4o achieves the top scores across all datasets, with average scores such as $0.656$ on HotpotQA and $0.814$ on 2WikiMultiHopQA, compared to $0.630$ and $0.734$ for GPT-4o-mini, respectively. Despite its reduced size, GPT-4o-mini remains highly competitive, frequently outperforming strong baselines like Search-o1. For instance, on 2WikiMultiHopQA, it achieves a Cover-EM score of $0.728$, closely matching or exceeding results from larger, task-specialized models. These findings demonstrate that our modular pipeline generalizes robustly across model scales, delivering strong multi-hop QA performance even under constrained computational budgets.

\begin{table}[ht]
\centering
\small
\caption{Comparison with prior work on Multi-Hop QA benchmarks. Each pair of values shows LLM Eval (left) and Cover-EM (right). All values are reported from the original papers. Evaluation settings may differ in aspects such as sample selection, prompt format, retriever, and passage truncation.}
\label{tab:scaling-benchmark}
\begin{tabular}{llcccc}
\toprule
\textbf{Method} & HotpotQA & 2WikiMultiHopQA & MuSiQue & Bamboogle & Avg\\
\midrule
\textbf{ReARTeR} \cite{sun2025rearter} & (0.506 / 0.468) & (0.534 / 0.554) & (0.302 / 0.296) & (0.544 / 0.496) & 0.463 \\
\textbf{R1-Searcher} \cite{song2025r1} & (\textbf{0.750} / \textbf{0.654}) & (0.650 / 0.636) & (0.314 / 0.282) & (0.544 / 0.528) & 0.545\\
\textbf{\OURS} & (0.702 / 0.558) & (\textbf{0.740} / \textbf{0.728}) & (\textbf{0.376} / \textbf{0.328}) & (\textbf{0.712} / \textbf{0.640}) & \textbf{0.598} \\
\bottomrule
\end{tabular}
\end{table}

\subsubsection{Comparison with other Prior Works}
To further contextualize the performance of \OURS under the GPT-4o-mini setting, we compare against retrieval-augmented baselines in \tref{tab:scaling-benchmark}. \textbf{ReARTeR} \cite{sun2025rearter} uses GPT-4o-mini as its generator, while \textbf{R1-Searcher} \cite{song2025r1} employs an RL-enhanced Qwen-2.5-7B. All values reflect the best results reported in their original papers. As shown, \OURS achieves the highest LLM Eval and Cover-EM scores on three of the four benchmarks—2WikiMultiHopQA, MuSiQue, and Bamboogle—trailing only R1-Searcher on HotpotQA. These findings highlight the strength of our modular design: it enables general-purpose models to perform competitively or even outperform fine-tuned, task-specific alternatives, despite using lighter model configurations.

\section{Discussion and Ablation}
\label{ablation}

To rigorously evaluate the efficacy of our modular design and the contributions of its constituent components, we conduct a series of ablation studies. These studies are performed on a randomly sampled subset of 200 examples from both the HotpotQA and MuSiQue datasets. For these ablation experiments, GPT-4o-mini is primarily employed as the base language model to ensure consistent and controlled comparisons.

\subsection{Effect of Self-Reflection on Multi-Hop QA Performance}
\label{ablation: self}
\begin{figure}[ht]
  \centering
  \includegraphics[width=\textwidth]{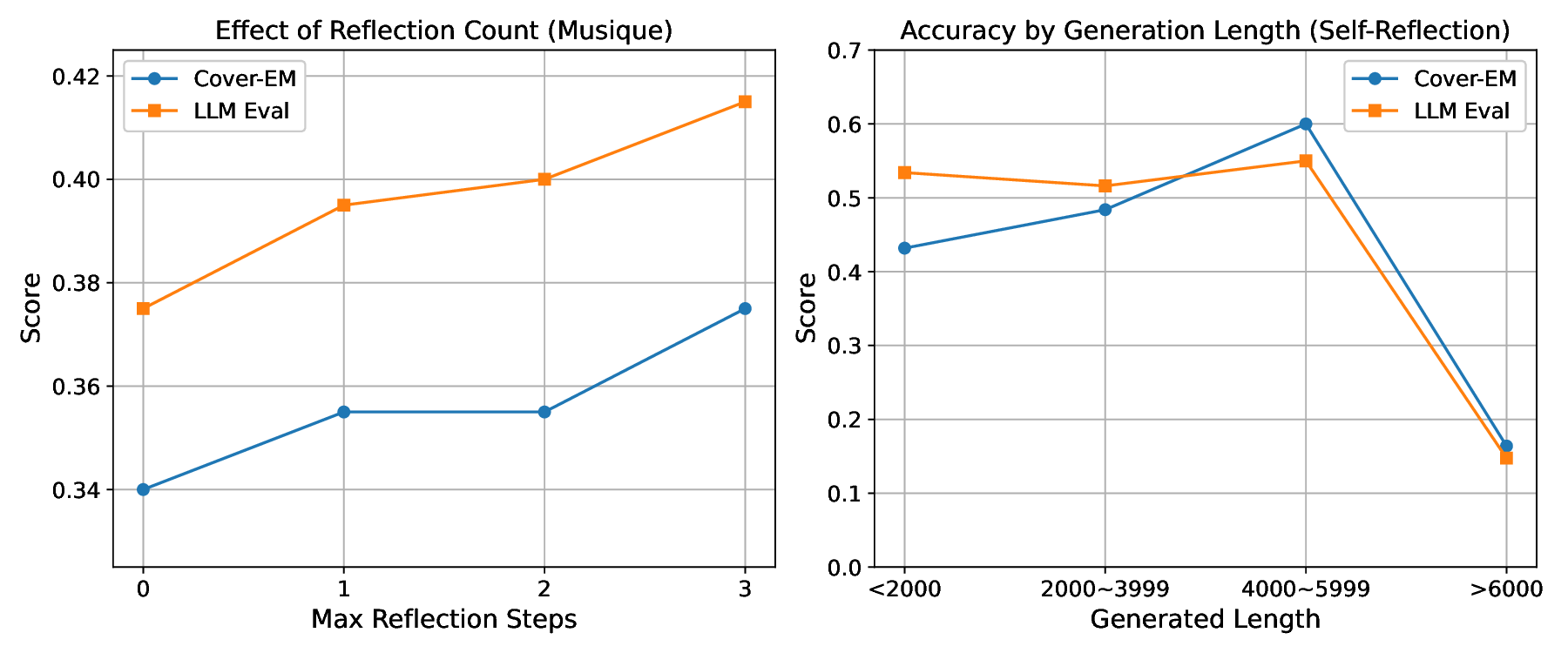}
  \caption{
    Accuracy of the self-reflection mechanism on the Musique dataset.
    \textbf{Left:} Accuracy improves as the number of reflection steps increases.
    \textbf{Right:} Performance by total generation length.
  }
  \label{fig:self-reflection}
\end{figure}
To enhance the robustness of multi-hop reasoning, \OURS incorporates a self-reflection mechanism. This feature permits the system to iteratively re-attempt the reasoning process up to a maximum of three times if the initial answer fails verification. The reflection process terminates early if a high-confidence answer is successfully verified. A configuration with zero reflection steps serves as the baseline, representing either a single pass through the multi-hop pipeline or an early exit via the Simple QA pathway.

As illustrated in Figure \ref{fig:self-reflection} (Left), the iterative self-reflection mechanism yields significant performance improvements on the Musique dataset. With each incremental reflection step, both Cover-EM and LLM Eval scores exhibit a steady increase. Specifically, Cover-EM rises from $0.340$ (at 0 reflections) to $0.375$ (at 3 reflections), while the LLM Eval score improves from $0.375$ to $0.415$ over the same range. This progression confirms that revisiting and refining earlier stages of the reasoning process enables the model to effectively correct initial errors and enhance the quality of its answers.

Figure \ref{fig:self-reflection} (Right) analyzes performance as a function of the total generation length, aggregated across all reflection steps. Optimal accuracy is observed for generations within the 4000–5999 token range, suggesting that moderately lengthy reflective reasoning is most efficacious. Conversely, a sharp decline in performance is noted for outputs exceeding 6000 tokens. This decline is likely attributable to unresolved complexities in the underlying questions or instances of reasoning drift during prolonged iterative processing, rather than an inherent flaw in the reflection mechanism itself.

In summary, these results demonstrate that bounded self-reflection is a valuable technique for improving multi-hop QA performance by facilitating iterative error correction. While extremely long generation traces correlate with lower accuracy, this primarily reflects the intrinsic difficulty of those specific queries.

\subsection{Ablation: Assessing the Contribution of Individual Modules}
\label{ablation: add}
\begin{table}[ht]
  \centering
  \small
  \caption{Module effectiveness for Multi-Hop QA}
  \label{tab:ablation}
  \begin{tabular}{cccccc|cc|cc}
    \toprule
    \multicolumn{6}{c|}{\textbf{Modules}} & \multicolumn{2}{c|}{\textbf{Hotpot-200}} & \multicolumn{2}{c}{\textbf{Musique-200}} \\
    QD & QC & QR & PR & AV & & Cover-EM & LLM Eval & Cover-EM & LLM Eval \\
    \midrule
    \xmark & \xmark & \xmark & \xmark & \xmark & & 0.480 & 0.625 & 0.170 & 0.195 \\
    \cmark & \cmark & \xmark & \xmark & \xmark & & 0.505 & 0.650 & 0.320 & 0.360 \\
    \cmark & \cmark & \xmark & \cmark & \xmark & & 0.510 & 0.665 & 0.335 & 0.365 \\
    \cmark & \cmark & \cmark & \xmark & \cmark & & 0.510 & 0.660 & 0.330 & 0.365 \\
    \cmark & \cmark & \cmark & \cmark & \cmark & & 0.505 & 0.680 & 0.350 & 0.375 \\
    \bottomrule
  \end{tabular}
\end{table}

To empirically validate the contribution of individual components within our modular architecture, as outlined in our key contributions (Section 1), we conducted an ablation study assessing the incremental impact of five core modules: Question Decomposition (QD), Question Construction (QC), Passage Reranking (PR), Query Rewriting (QR), and Answer Verification (AV). The results of this stepwise module integration, presented in \tref{tab:ablation}, are measured using both Cover-EM and LLM-based evaluation. It is important to note certain inherent couplings in module functionality: QD and QC are necessarily paired, as decomposition generates sub-questions that require subsequent construction into executable queries. Similarly, QR is typically activated following verification failures, making it closely integrated with AV. Our evaluation strategy respects these dependencies.

The analysis commences with a baseline configuration where no specialized modules are active, resulting in limited performance, particularly on the more complex MuSiQue dataset ($0.170$ Cover-EM, $0.195$ LLM Eval). The introduction of the foundational QD and QC modules yields the most substantial performance improvements. For instance, on MuSiQue, this addition nearly doubles the Cover-EM to $0.320$ and significantly boosts the LLM Eval score to $0.360$, underscoring the critical role of structured decomposition and sub-question formulation in multi-hop reasoning. The subsequent integration of the PR module, which enhances context quality through improved evidence ranking, provides further incremental gains (e.g., on HotpotQA, LLM Eval increases from $0.650$ to $0.665$ when PR is added to the QD and QC configuration). Finally, the inclusion of the QR and AV modules, which refine intermediate reasoning steps and filter unsupported answers, contributes to further enhancements, particularly in the LLM evaluation metric (e.g., on MuSiQue, LLM Eval rises from $0.365$ with QD, QC, and PR to $0.375$ when all modules are active). These findings collectively confirm that each module provides a distinct and valuable contribution to the overall pipeline performance. The initial decomposition and construction modules establish a strong foundation, while subsequent modules for reranking, rewriting, and verification progressively refine the reasoning process, enhancing answer precision and robustness.

\subsection{Upgrade: Independent Contribution of Upgraded Modules}
\label{ablation: upgrade}
\begin{table}[ht]
\centering
\small
\caption{Impact of individually upgrading modules}
\label{tab:module-upgrades}
\begin{tabular}{ccccccc|cc}
\toprule
QD & QC & QR & PR & AG & AV & & Cover-EM & LLM Eval \\
\midrule
\ocircle & \ocircle & \ocircle & \ocircle & \ocircle & \ocircle & & 0.350 & 0.375 \\
\cmark & \ocircle & \ocircle & \ocircle & \ocircle & \ocircle & & 0.350 & 0.390 \\
\cmark & \cmark & \ocircle & \ocircle & \ocircle & \ocircle & & 0.355 & 0.395 \\
\cmark & \cmark & \cmark & \ocircle & \ocircle & \ocircle & & 0.355 & 0.405 \\
\cmark & \cmark & \cmark & \cmark & \ocircle & \ocircle & & 0.380 & 0.420 \\
\cmark & \cmark & \cmark & \cmark & \cmark & \ocircle & & 0.380 & 0.430 \\
\cmark & \cmark & \cmark & \cmark & \cmark & \cmark & & 0.385 & 0.450 \\
\bottomrule
\end{tabular}
\end{table}
A key tenet of \OURS, as highlighted in the Introduction, is the independent upgradability of its modules, contributing to the framework's flexibility and scalability. To empirically demonstrate this, we investigate the standalone benefit of enhancing individual modules by comparing two operational versions: a baseline configuration where each module is implemented using \texttt{gpt-4o-mini} (denoted as \ocircle), and an enhanced version where a specific module is upgraded to use the more powerful \texttt{gpt-4o} model (denoted as \cmark). The Answer Generation (AG) module, a constant presence in the pipeline, is also subject to this upgradability. These experiments were conducted on the Musique-200 subset.

The study begins with a baseline where all modules utilize \texttt{gpt-4o-mini}, establishing an LLM Eval score of $0.375$. We then progressively upgrade one module at a time to its \texttt{gpt-4o} counterpart. As detailed in \tref{tab:module-upgrades}, upgrading the Question Decomposition (QD) module alone results in an immediate increase in the LLM Eval score from $0.375$ to $0.390$. Subsequent individual upgrades to other modules, such as Question Construction (QC), Query Rewriting (QR), Passage Reranking (PR), Answer Generation (AG), and Answer Verification (AV), each contribute to further, steady improvements in performance. When all modules are upgraded to utilize \texttt{gpt-4o}, the LLM Eval score reaches $0.450$. These results clearly confirm that enhancing individual modules in isolation yields measurable and additive performance gains. This validates the independent upgradability feature of \OURS, underscoring its capacity to adapt to varying computational resources and to benefit from advancements in individual model capabilities without requiring a complete overhaul of the pipeline. This inherent flexibility and scalability are central to the practical utility of our framework.

\subsection{Efficiency-Oriented Design: Retrieval Decision and Simple QA Handling}
\label{ablation: efficiency}
\begin{table}[ht]
\centering
\small
\caption{Impact of Retrieval Decision on Hotpot-200}
\label{tab:retrieval-decision}
\begin{tabular}{lcc}
\toprule
\textbf{Decision} & \textbf{Cover-EM} & \textbf{LLM Eval} \\
\midrule
no & 0.540 & 0.725 \\
gpt-4o-mini & 0.535 & 0.745 \\
gpt-4o & 0.555 & 0.745 \\
\bottomrule
\end{tabular}
\end{table}

\begin{table}[ht]
\centering
\small
\caption{Impact of adding Simple QA on Hotpot-200 (\texttt{gpt-4o-mini})}
\label{tab:simpleqa}
\begin{tabular}{lccc}
\toprule
\textbf{Simple QA} & \textbf{Cover-EM} & \textbf{LLM Eval} & \textbf{Avg. Tokens} \\
\midrule
no & 0.505 & 0.680 & 1219 \\
yes & 0.530 & 0.695 & 832 \\
\bottomrule
\end{tabular}
\end{table}

To bolster computational efficiency without unduly compromising accuracy, \OURS incorporates two dynamic components: the Retrieval Decision module and a dedicated Simple QA pipeline. The Retrieval Decision module intelligently determines whether external knowledge retrieval is necessary for a given (sub-)question. This allows the system to bypass the retrieval step if the input context already contains sufficient information for grounded reasoning, thereby conserving resources. It is important to note that this decision does not imply a fallback to ungrounded parametric knowledge; rather, it prioritizes leveraging already available, contextualized information. Complementing this, the Simple QA module is designed to identify and directly process questions that do not require complex multi-hop reasoning. By routing such "easy" questions through a lightweight path, it avoids the computational overhead associated with the full multi-hop pipeline.

The efficacy of these efficiency-oriented components was evaluated on the Hotpot-200 subset, using both \texttt{gpt-4o-mini} and \texttt{gpt-4o} as the backbone LLM. As indicated in \tref{tab:retrieval-decision}, activating the Retrieval Decision module (comparing "no" decision to using \texttt{gpt-4o-mini} or \texttt{gpt-4o} for the decision) generally maintains or slightly improves performance. For instance, with \texttt{gpt-4o-mini}, the LLM Eval score increases from $0.725$ (no decision module) to $0.745$ when the decision module is active. The optimal configuration, employing \texttt{gpt-4o} for the decision, achieves the highest Cover-EM of 0.555 while matching the top LLM Eval score of $0.745$.

Furthermore, \tref{tab:simpleqa} demonstrates the benefits of integrating the Simple QA pipeline. When this pathway is enabled for the Hotpot-200 subset (using \texttt{gpt-4o-mini}), there is a marked reduction in average token consumption per query, decreasing by approximately $32\%$ (from 1219 to 832 tokens). Concurrently, both Cover-EM (from $0.505$ to $0.530$) and LLM Eval (from $0.680$ to $0.695$) scores exhibit improvement. These results collectively underscore the advantages of \OURS's flexible, modular design, where adaptive routing mechanisms like early exits for simple questions and conditional retrieval can be seamlessly integrated to optimize for both efficiency and effectiveness.seamlessly integrated to improve efficiency without compromising effectiveness.

\subsection{Impact of Generation Length on Accuracy}
\label{ablation: generation}
\begin{figure}[ht]
  \centering
  \includegraphics[width=\textwidth]{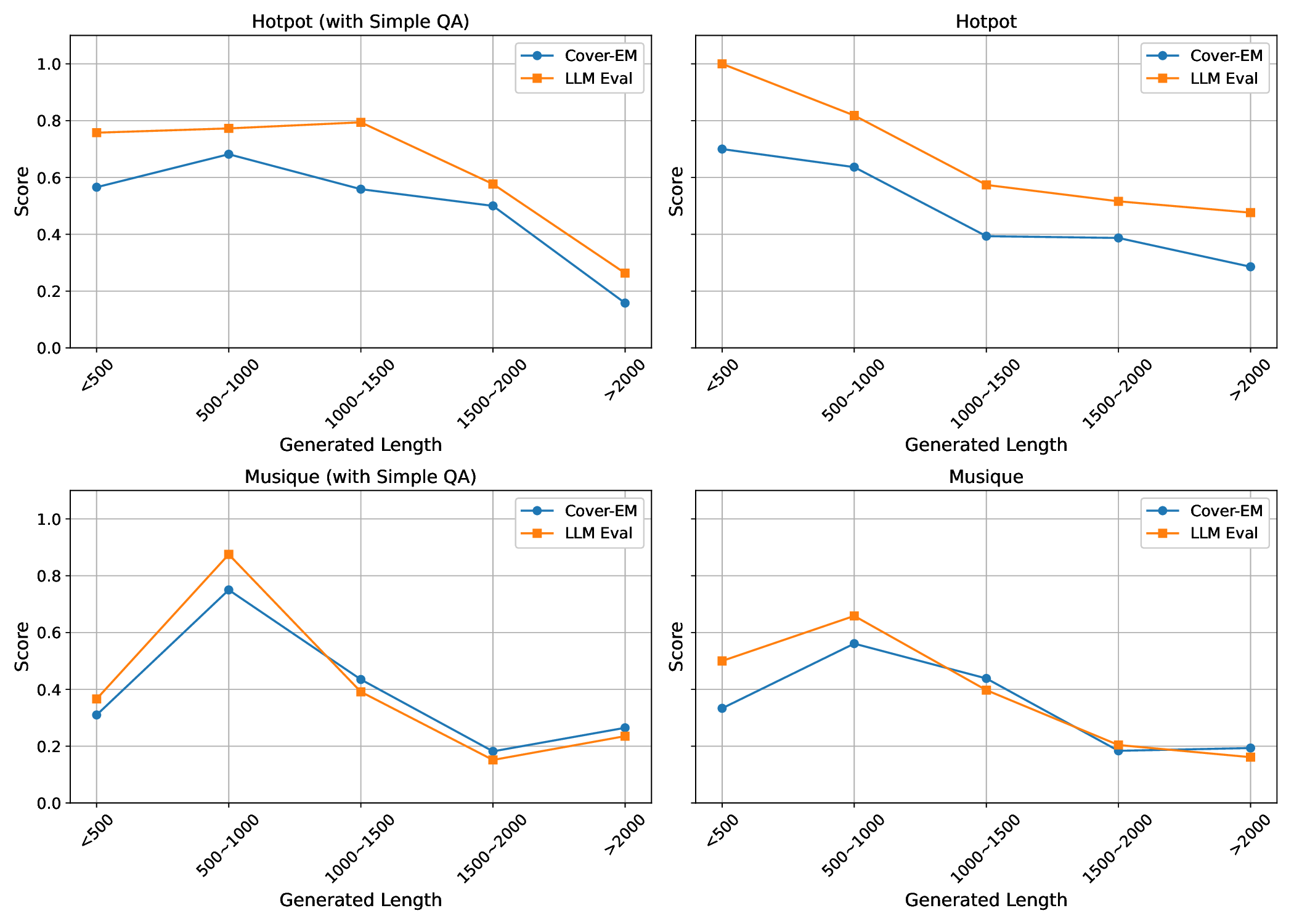}
  \caption{
    Accuracy across different generation length intervals. Each subplot shows the relationship between output length and performance for Hotpot and Musique.
  }
  \label{fig:length-vs-acc}
\end{figure}

To investigate the relationship between the length of generated outputs and system performance, we conducted an analysis where model outputs were categorized based on their total generated token count. These generations were divided into five distinct length intervals. Within each interval, we compared Cover-EM and LLM evaluation scores for two configurations of our system: the standard Multi-Hop QA pipeline and an augmented version incorporating the Simple QA pipeline.

The results, depicted in Figure \ref{fig:length-vs-acc}, reveal a general trend where longer generations are associated with lower accuracy. Specifically for the HotpotQA dataset, the Cover-EM score decreases markedly from $0.700$ for outputs containing fewer than 500 tokens to a mere $0.286$ for those exceeding 2000 tokens. This suggests that extended outputs often correspond to more complex or unresolved queries, potentially reflecting a more diffuse or less successful reasoning process. A similar pattern is observable for the MuSiQue dataset, although the relationship appears more nuanced, likely due to the inherently higher difficulty of this benchmark. In this context, very short generations might indicate premature termination of the reasoning process, whereas outputs of moderate length tend to correlate with optimal performance.

The introduction of the Simple QA pipeline demonstrates a positive impact on performance for mid-length generations, primarily by enabling efficient early exits for questions that are solvable without extensive multi-hop reasoning. However, a slight decrease in accuracy is noted for shorter generations on the MuSiQue dataset when the Simple QA pipeline is active. This may suggest instances where the system occasionally misclassifies complex queries as simple, leading to underperformance due to premature routing.

Overall, this analysis highlights the intricate connection between generation length, the inherent difficulty of the input question, and the behavior of the pipeline. Monitoring the length of generated outputs can therefore serve as a valuable heuristic for refining routing decisions and verification strategies within modular question answering systems like \OURS.
\section{Conclusion}
In this work, we presented a modular abstraction for Multi-Hop RAG that addresses the limitations of existing monolithic QA pipelines. By decomposing the reasoning process into a sequence of parameterized, composable modules, our framework promotes interpretability, supports component-wise analysis, and enables flexible reconfiguration for diverse tasks. Across four multi-hop QA benchmarks, our modular pipeline consistently outperforms strong baselines like RQ-RAG and Search-o1, while maintaining scalability across large language models of varying capacities. Notably, our system enhances factual grounding through verification and abstention mechanisms, and recovers from errors via iterative self-reflection. These results demonstrate that a modular design can match or exceed the performance of specialized systems while offering greater transparency, adaptability, and extensibility for future RAG development.

\section{Limitations}
Despite its strengths, our approach has several limitations. First, the performance of the overall system is tightly coupled to the reliability of individual modules, particularly the Question Decomposition module. When questions exhibit subtle logic or ambiguous structure, decomposition may still fail, even with self-reflective refinement. Second, our method relies on large instruction-tuned LLMs for module execution, which introduces latency and resource overhead. This may limit the framework's applicability in low-resource or real-time settings. Future work may explore lighter-weight alternatives or dynamic adaptation strategies to reduce inference cost while preserving modular reasoning quality.

\bibliographystyle{plainnat}
\bibliography{main}

\clearpage
\appendix
\section{Supplementary Modules and Pipeline Details}
\subsection{Detailed Modular Components}
\label{appendix:modules}
\paragraph{Question Decomposition}
Multi-hop questions often require synthesizing information spread across multiple documents or reasoning steps. To handle such complexity, we introduce this module that leverages a large language model to break down the original question into a sequence of interrelated sub-questions. Unlike prior approaches that structure reasoning through a tree \cite{zhang2023reasoning} or graph-based decomposition \cite{verma2024plan}, our method follows the format proposed in \cite{trivedi2022musique}, prompting the model to generate sub-questions sequentially. This linear, interconnected formulation encourages the model to uncover and follow the underlying logical flow of the original question, ensuring each sub-question builds on previously retrieved or inferred information. Given an input question \( q \in \mathcal{Q} \), the goal of the decomposition module is to generate two components: a set of sub-questions \( \mathbf{S} = \{s_1, s_2, \dots, s_n\} \subset \mathcal{Q} \), where each \( s_i \) represents a single-hop question derived from \( q \); and a reasoning trace \( r \in \mathcal{R} \), which provides the reasoning path needed to decompose \( q \). Here, \( \mathcal{Q} \) represents the space of all questions and \( \mathcal{R} \) represents the space of all reasoning traces. Formally, we define the decomposition process as: \((\mathbf{S}, r) = \mathrm{Decompose}_{\theta}(q)\), where \( \mathrm{Decompose}_{\theta} \) is a function parameterized by the LLM weights \( \theta \). This module enables downstream modules to independently retrieve evidence and give the answer.

\paragraph{Question Construction}
This is designed to rewrite an abstract or context-dependent sub-question into a fully specified, self-contained form. This is especially important in multi-hop question answering, where a sub-question may refer to previous answers (e.g., using placeholders like "\#1" or pronouns like "he" or "it"). This module takes the following inputs:
\begin{itemize}
    \item \textbf{Original Question} \( q \in \mathcal{Q} \): The full multi-hop question that provides overall context.
    \item \textbf{Previous QA Pairs} \( H = \{(s_1, r_1, a_1), \dots, (s_{i-1}, r_{i-1}, a_{i-1})\} \): The history of sub-questions and their corresponding answers with reasoning from earlier steps.
    \item \textbf{Current Sub-question} \( s_i \in \mathcal{Q} \): The context-dependent sub-question that may contain references to earlier answers.
\end{itemize}
Using these inputs, the model constructs \textbf{Fully-Specified Sub-question} \( \hat{s}_i \in \mathcal{Q} \): a rewritten version of \( s_i \) where all references are resolved based on the previous answers. So the construction process is defined as: \(\hat{s}_i = \mathrm{Construct}_\theta(q, H, s_i)\), where \( \mathrm{Construct}_\theta \) is a function parameterized by the LLM weights \( \theta \), mapping the original question \( q \), previous QA history \( H \), and current sub-question \( s_i \) to the resolved sub-question \( \hat{s}_i \). This module ensures that all sub-questions are well-formed, explicit, and interpretable.

\paragraph{Retrieval Decision}
This module determines whether external retrieval is necessary for answering a given question. In some cases, the question already contains sufficient information for making a decision, such as comparison, reasoning, or arithmetic over known entities. By avoiding unnecessary retrievals, the system can reduce latency and improve efficiency while maintaining answer quality.

The retrieval decision process function is like: \((d, r) = \mathrm{Decide}_{\theta}(q)\), where \( q \in \mathcal{Q} \) is the input question, \( d \in \{\texttt{True}, \texttt{False}\} \) is a binary decision indicating whether retrieval is needed, and \( r \in \mathcal{R} \) is a reasoning trace justifying the decision. When \( d = \texttt{False} \), the pipeline proceeds directly to reasoning or answering modules without invoking retrieval, otherwise will enable rertieval for the current question answering.
\paragraph{Query Rewriting}
In order to iteratively refine the search query to improve retrieval relevance and alignment with the user’s intent, we have one module for query rewriting. At each step, it takes the original question and the last rewritten version of the query and generates a new, improved query that better captures the current information need. This approach draws inspiration from prior work \cite{chan2024rq}, which demonstrates the effectiveness of intermediate query reformulation in multi-hop retrieval.

We define the query rewriting process as a function:
\(q' = \mathrm{Rewrite}_\theta(q, q_{\text{prev}})\), where \( q \in \mathcal{Q} \) is the original question, \( q_{\text{prev}} \in \mathcal{Q} \) is the previous version of the rewritten query, and \( q' \in \mathcal{Q} \) is the new query output. The function \( \mathrm{Rewrite}_\theta \) is implemented by a LLM with parameters \( \theta \), mapping the current and previous queries to an improved query formulation.

\paragraph{Passage Reranking}
Inspired by \cite{sun2023chatgpt}, this module aims to reorder a set of retrieved passages based on their relevance to a given search query. The reranking function considers both the semantic alignment between the query and each passage, and the informativeness of the content. Importantly, it also helps mitigate the lost-in-the-middle problem by promoting highly relevant passages to the top of the ranked list, since relevant evidence may be present but overlooked due to suboptimal ordering.
We define the passage reranking process as: \((r, \pi) = \mathrm{Rerank}_\theta(q, I, P)\), where \( q \in \mathcal{Q} \) is the input search query, which may be either the original full question or a rewritten version of it. The set \( I = \{i_1, i_2, \dots, i_k\} \) contains the numerical identifiers corresponding to the retrieved passages \( P = \{p_1, p_2, \dots, p_k\} \). The function \( \mathrm{Rerank}_\theta \), parameterized by LLM weights \( \theta \), returns a reasoning trace \( r \in \mathcal{R} \) and a permutation \( \pi \in \Pi \) over the identifiers \( I \), representing the final ranked order. After post-processing for the output ranking order, we can get the full reranked passages for future use. 
The reranking plays a crucial role in ensuring that the most relevant evidence is prioritized for downstream reasoning and generation.

\paragraph{Answer Generation}
The Answer Generation module is responsible for producing the final answer to a given question by leveraging retrieved evidence and background information. It is designed to ensure that the answer is both accurate and grounded in the supporting context.
This module takes three inputs:
\begin{itemize}
    \item \textbf{Question} \( q \): the current question, which may be either the original full question or a sub-question generated through decomposition.
    \item \textbf{Retrieved Passages} \( P = \{p_1, p_2, \dots, p_k\} \): evidence passages retrieved from the knowledge corpus.
    \item \textbf{Background} \( b \): contextual information to support reasoning. For an original full question, the background is set to "none". For a decomposed sub-question, the background includes prior sub-questions and their answers, helping to disambiguate and resolve incomplete context from earlier hops.
\end{itemize}
The output includes a reasoning trace \( r \in \mathcal{R} \), which is an explanation that justifies the answer based on the input context and retrieved content; and an answer with citation \( a \in \mathcal{A} \), which is a factual answer explicitly supported by the retrieved passages and includes citation markers referring to the evidence.
We define the answer generation process as: \((a, r) = \mathrm{AnswerGen}_\theta(q, P, b)\), where \( \mathrm{AnswerGen}_\theta \) is a generation function parameterized by the LLM weights \( \theta \), mapping the tuple \( (q, P, b) \) into the answer and its corresponding reasoning trace.

This setup enables the model to support both one-hop and multi-hop scenarios. In multi-hop settings, the background input \( b \) plays a critical role in maintaining consistency and grounding across steps, especially when earlier answers provide essential but non-retrievable information.

\paragraph{Answer Verification}
This module evaluates whether a generated answer is both (1) factually correct with respect to the original question, and (2) properly grounded in the retrieved support passages.
The answer verification process is defined as a function: \((y, r) = \mathrm{Verify}_\theta(q, a, P)\), where \( q \in \mathcal{Q} \) is the input question, \( a \in \mathcal{A} \) is the generated answer, and \( P = \{p_1, p_2, \dots, p_k\} \) is the set of retrieved support passages. The output \( y \in \{\texttt{True}, \texttt{False}\} \) is a binary label indicating whether the answer is valid, and \( r \in \mathcal{R} \) is a reasoning trace justifying the verification outcome. The function \( \mathrm{Verify}_\theta \) is implemented by a LLM parameterized by weights \( \theta \). This module serves to filter out hallucinated or unsupported answers, and enforces strict grounding by requiring that every verified answer be explicitly justified using the retrieved content.

\paragraph{Final Answering}
This module is responsible for synthesizing the final answer to the original multi-hop question by integrating all intermediate results from the reasoning process. It takes as input the original question and an answer history, which includes the sequence of sub-questions, their corresponding reasoning traces, and their answers.

We define the final answering process as: \((a, r) = \mathrm{Finalize}_\theta(q, H)\), where \( q \in \mathcal{Q} \) is the original question, \( H = \{(s_1, r_1, a_1), \dots, (s_n, r_n, a_n)\} \) is the answer history consisting of sub-questions \( s_i \), reasoning traces \( r_i \), and answers \( a_i \), and \( \mathrm{Finalize}_\theta \) is a function parameterized by LLM weights \( \theta \). The output is the final answer \( a \in \mathcal{A} \) and a comprehensive reasoning trace \( r \in \mathcal{R} \) that explains the aggregation of intermediate steps.

This step completes the multi-hop pipeline by assembling fragmented insights into a single, well-justified final prediction.

\subsection{Simple QA pipeline}
\label{appendix:simple}
This is designed to answer straightforward questions that can be resolved using a single retrieval step. First, the system retrieves a set of relevant passages \( P = \{p_1, p_2, \dots, p_k\} \) from the document corpus using the retrieval function: \(P = \mathrm{Retrieve}(q)\). These passages, along with the question \( q \), are passed to the \textbf{Answer Generation} module, which uses a language model to produce an answer \( a \in \mathcal{A} \) and a reasoning trace \( r \in \mathcal{R} \). Since the model will use the whole question to answer, the background input is set to \texttt{none}: \((a, r) = \mathrm{AnswerGen}_\theta(q, P, \texttt{none})\). The candidate answer \( a \) is then validated by the \textbf{Answer Verification} module, which determines whether the answer is correct and grounded in the retrieved passages: \((y, r_v) = \mathrm{Verify}_\theta(q, a, P)\). Here, \( y \in \{\texttt{True}, \texttt{False}\} \) is a binary decision, and \( r_v \in \mathcal{R} \) is a verification reasoning trace. Based on the verification result, the final output is determined as:
\begin{equation}
\text{FinalAnswer} =
\begin{cases}
a, & \text{if } y = \texttt{True} \\
\texttt{Answer from Multi-Hop QA pipeline}, & \text{if } y = \texttt{False}
\end{cases}
\end{equation}
If the answer is verified as correct, it is returned as the final output. Otherwise, the pipeline defers to a more advanced Multi-Hop QA pipeline for further processing. Simple QA pipeline ensures both efficiency and reliability by quickly resolving simple questions, but it encounters failures like retrieving only partial evidence from a single query or failing to identify implicit logical structure within the question. To address this, the question decomposition and construction design gives Multi-Hop QA Pipeline more advancement, which explicitly breaks the original question into a sequence of sub-questions and solve them step-by-step.

\section{Compatibility with Pipeline Design and Fair Comparison}
\label{appendix:comparison}
For RQ-RAG, we use the fine-tuned llama2-7b model as in its original implementation. However, our system is designed for prompt-based modular reasoning without task-specific fine-tuning. We use Llama3.1-8b instead of llama2-7b for two reasons: (1) llama2-7b struggles to follow our structured prompts without fine-tuning, often yielding unparseable outputs; (2) Llama3.1-8b provides stronger instruction-following out of the box, making it more compatible with our pipeline. This reflects a broader contrast: RQ-RAG uses a fine-tuned small model, while our system demonstrates how general-purpose, instruction-tuned models can perform well via simple prompting, without any fine-tuning.

For Search-o1, we retain QwQ-32B, a reasoning-specialized model designed for multi-step inference and self-reflective answering. Replacing it with a general-purpose LLM would undermine its intended strengths. In contrast, our pipeline is designed to be model-agnostic and works best with instruction-following LLMs rather than reasoning-specialized engines. To ensure a meaningful comparison, we evaluate our system using a range of general-purpose LLMs chosen for both performance and compatibility with our prompt-based design. In particular, we use Qwen2.5-72B-Instruct, a large instruction-tuned model chosen to compare fairly with QwQ-32B, since reasoning-focused models are typically more efficient and capable at reasoning despite having fewer parameters. Moreover, we use GPT-4o to demonstrate how our pipeline scales with a more powerful model, highlighting its ability to fully leverage the capabilities of advanced LLMs. In addition, we include GPT-4o-mini, a significantly smaller variant, to show that our system maintains strong performance even with limited model capacity, reinforcing the pipeline’s efficiency, robustness, and adaptability across a wide range of compute budgets.

Importantly, we carefully control for retrieval to ensure fairness. RQ-RAG and our method both use the wiki2018 corpus when compared directly, as RQ-RAG is fine-tuned using retrieved knowledge from this version. In comparisons involving Search-o1, we use wiki2019 for both Search-o1 and our method, as it contains more up-to-date information, reducing bottlenecks in the retrieval stage and allowing both methods to demonstrate their full capabilities. Additionally, we fix the retriever to Cortex Search for both our system and Search-o1 to ensure consistent retrieval quality, while RQ-RAG uses its original e5-base-v2 retriever as part of its design.

\section{Grouding Analysis}
\label{appendix:grounding}
\subsection{Representative cases in wiki2019}
Several representative cases illustrate the grounding issue. For example, in response to the question \textit{"Who was born later, \textbf{Meinhart Maur} or \textbf{Slaviča Kundačina}?"}, the retrieval contains the birth date of \textit{Meinhart Maur} but provides no information about \textit{Slaviča Kundačina}. However, the model identifies a similar name—\textit{Slavica Ćukteraš}, which assumes they are the same person, answering based on that speculative link. In another case, \textit{"Who is the father of the director of the film Les Truands?"}, the director is \textit{Carlo Rim}, whose real name is \textit{Jean Marius Richard}. Although no retrieved document directly names his father, the model infers it to be \textit{Marius Richard}, reasoning from the director's birth name. Similarly, for \textit{"What is the date of birth of the editor of The Texas Ranger (Magazine)?"}, the retrieved context fails to identify the editor, but the model uses its internal knowledge to answer the candidate and then provides the corresponding birth date. 

\subsection{More detailed analysis in wiki2018}
During the experiments on 2WikiMultiHopQA dataset, we found that many questions cannot be answered with the wiki2018 corpus due to lack of knowledge. We present the performance of our system and the Search-o1 baseline using wiki2018 in \tref{2018-table}. Among the examples that Search-o1 can generate correct answers but we give "I don't know." based on grounding consideration. We found the following issues which are summarized in \tref{tab:grounding}.
\paragraph{Assumption under insufficient information}
When the retrieved information is incomplete or ambiguous, the reasoning model may generate answers by making implicit assumptions. This behavior can lead to plausible-sounding but ungrounded responses. For example, consider the question: \textit{"Which film has the director who died first, \textbf{Convoy Buddies} or \textbf{Tip Toes}?"} In the retrieved passages, no explicit information is available regarding the director of \textit{Convoy Buddies}. However, the passages mention that \textit{Paul L. Smith} starred in the film. The reasoning model incorrectly assumes that \textit{Paul L. Smith} is the director and proceeds to retrieve and compare his death date \textit{April 25, 2012} with that of the director of \textit{Tip Toes}. While this produces a definitive answer, the conclusion is unsupported and results from speculative reasoning due to the lack of direct evidence.

\paragraph{Misidentification under semantic similarity}
The reasoning model may mistakenly align a query entity with a semantically similar but incorrect one when the exact match is not found in the retrieved content. For example, consider the question: \textit{"Which album was released first, \textbf{L.A. To Miami} or \textbf{Today Is Our Valentine's Day}?"} In this case, there is no album or song with the exact title \textit{Today Is Our Valentine's Day} in the corpus. However, the retrieved passages include information about the 2010 soundtrack album \textit{Valentine's Day}, which was released on \textit{February 9, 2010}. The model incorrectly interprets this as referring to the target entity and uses its release date to make a comparative judgment.

\paragraph{Factual hallucination without retrieval}
In some cases, the reasoning model produces factual claims that are not grounded in any retrieved evidence, instead relying on its internal parametric knowledge, which may be outdated or incorrect. For example, consider the question: \textit{"Which film has the director who died earlier, \textbf{La Moglie Vergine} or \textbf{Melvin and Howard}?"} In this case, the model fails to initiate retrieval for information about the director of \textit{Melvin and Howard}. Instead, it hallucinates an answer by incorrectly asserting that the director is \textit{Michael Ritchie} and proceeds to make a comparison based on this unsupported claim.

\paragraph{Default assumption based on common knowledge}
Sometimes the reasoning model may default to assumptions based on general world knowledge or statistical priors. For example, consider the question: \textit{"Who was born later, \textbf{Bradley Tyler Johnson} or \textbf{Bernardo Freitas}?"} In the absence of retrieved birthdate information for both individuals, the model responds by reasoning that many Brazilian athletes are typically born in the 1990s or later. Based on this assumption, it concludes that Bernardo is likely the younger individual. While this guess may seem plausible, it is not grounded in any retrieved facts.

\paragraph{Random selection under uncertainty}
In scenarios where the retrieved evidence does not provide sufficient information to support either option, the model may still attempt to produce an answer by making an arbitrary or speculative choice. For example, consider the question: \textit{"Who is younger, \textbf{Jan Britstra} or \textbf{Leonas Milčius}?"} In this case, there is no retrieved information containing the birthdates or ages of either individual. Despite the lack of evidence, the model responds with: \textit{"Since I have to choose, perhaps the intended answer is that Leonas Milčius is younger, so I'll go with that."} This behavior reflects an ungrounded decision made under uncertainty, where the model chooses an answer simply to avoid abstaining.

\paragraph{Attribution under insufficient information}
The reasoning model may attempt to attribute information by referencing partial or indirectly related mentions. For example, consider the question: \textit{"Which film has the director born later, \textbf{Cose Da Pazzi} or \textbf{Kadhalil Sodhappuvadhu Yeppadi}?"} In this case, no reliable birth-year information is found for the actual director of \textit{Cose Da Pazzi}. However, the model references possible candidates based on partial associations with the film or Italian cinema. It then infers that the director of the Italian film is likely older, and concludes that \textit{Kadhalil Sodhappuvadhu Yeppadi} has the younger director. This attribution, while seemingly reasonable, is based on indirect evidence.

\begin{table}
  \caption{Results table for 2wikimultihop dataset using wiki2018 corpus}
  \label{2018-table}
  \centering
  \begin{tabular}{llllll}
    \toprule
    \multicolumn{6}{l}{\textbf{2WikiMultiHopQA}} \\
    \midrule
    Method     & LLM     & Retriever     & Corpus     & LLM Eval     & Cover-EM     \\
    \midrule
    Search-o1 & QwQ-32B & Cortex Search & wiki2018 & 0.664 & 0.622 \\
    Ours & Qwen2.5-72B-Instruct & Cortex Search & wiki2018 & 0.484 & 0.502  \\
    Ours & GPT-4o & Cortex Search & wiki2018 & 0.574 & 0.564  \\
    \bottomrule
  \end{tabular}
\end{table}

\begin{table}
  \caption{Grounding issues in experiments}
  \label{tab:grounding}
  \centering
  \begin{tabular}{lll}
    \toprule
    & Problem     & Proportion \\
    \midrule
    1 & Assumption under insufficient information & 6.0\% \\
    2 & Misidentification under semantic similarity & 3.0\% \\
    3 & Factual hallucination without retrieval & 0.6\% \\
    4 & Default assumption based on common knowledge & 0.4\% \\
    5 & Random selection under uncertainty & 0.2\% \\
    6 & Attribution under insufficient information & 0.2\% \\
    \midrule
    & Total & 10.4\% \\
    \bottomrule
  \end{tabular}
\end{table}

\newpage
\section{LLM Prompts and Templates}
In this section, we present the instruction prompts and the input/output template provided to LLMs.

\begin{tcolorbox}[mypromptbox, title=Prompt for Question Decomposition - Example Part]
Question: What is the capital of France?\\
Reasoning: The question asks for a direct fact that does not require multiple steps or dependencies.\\
Output: None\\
Question: Who is the CEO of the company owns the brand that produces iPhones?\\
Reasoning: To answer this question, it is necessary to identify the company associated with the brand producing iPhones first. Once the company is determined, its CEO can be identified. This logical dependency makes decomposition necessary.\\
Output:\\
1. What is the company owns the brand that produces iPhones?\\
2. Who is the CEO of \#1?\\
Question: Which country has a larger population, Canada or Australia?\\
Reasoning: The question requires a comparative analysis. To perform the comparison, the populations of Canada and Australia must first be individually achieved. Then, the comparison can be made based on the given values.\\
Output:\\
1. What is the population of Canada?\\
2. What is the population of Australia?\\
3. Canada has a population of \#1. Australia has a population of \#2. Which country has a larger population, Canada or Australia?\\
Question: Which river is longer, Nile or Tigris?\\
Reasoning: This question requires determining the lengths of the two rivers to make a comparison. The lengths of each river must first be individually identified before a conclusion can be drawn.\\
Output:\\
1. What is the length of Nile?\\
2. What is the length of Tigris?\\
3. The length of Nile is \#1. The length of Tigris is \#2. Which river is longer, Nile or Tigris?\\
Question: What is the capital city of the country where the painter of Starry Night was born?\\
Reasoning: The question involves a chain of relationships. First, the painter of Starry Night must be identified. Then, the country where this painter was born needs to be determined. Finally, the capital city of that country must be retrieved. Each step depends on the outcome of the previous one, making decomposition necessary.\\
Output:\\
1. Who is the painter of Starry Night?\\
2. What is the country where \#1 was born?\\
3. What is the captital city of \#2?\\
Question: What is the distance between the tallest mountain in Japan and the tallest mountain in Nepal?\\
Reasoning: To find the distance, it is first essential to identify the tallest mountains in Japan and Nepal individually. Once both are identified, their geographical distance can be obtained. Decomposition ensures clarity and systematic problem-solving.\\
Output:\\
1. What is the tallest mountain in Japan?\\
2. What is the tallest mountain in Nepal?\\
3. What is the distance between \#1 and \#2?\\
Question: What is the distance between the city where the Colosseum is located and the capital city of the country where the Eiffel Tower is located?\\
Reasoning: This question involves two distinct entities and requires their geographical locations to be identified. First, the country where the Eiffel Tower is located must be found, followed by identifying its capital city. Next, the city where the Colosseum is located must be identified. Finally, the distance between these two locations can be determined. The multiple layers of reasoning and dependencies make decomposition necessary.\\
Output:\\
1. What is the country where the Eiffel Tower is located?\\
2. What is the capital city of \#1?\\
3. What is the city where the Colosseum is located?\\
4. What is the distance between \#3 and \#2?
\end{tcolorbox}

\begin{tcolorbox}[mypromptbox, title=Prompt and Template for Question Decomposition]
You are an expert skilled at analyzing complex questions and decomposing them into distinct and simpler sub-questions.\\
Your task is to analyze the input question and determine whether it requires decomposition.\\
Please give the reasoning that explains why decomposition was needed and the logic behind breaking the question into sub-questions.\\
If the question is already simple and does not require decomposition, output "None". Otherwise, decompose the question into multiple distinct, interconnected, self-contained sub-questions.\\
If a sub-question depends on the answer to a previous sub-question, refer to the previous sub-question using "\#number", don't use the "based on".\\
For questions that are simple, they do not require identifying multiple intermediate steps or facts before answering. They can be directly answered by retrieving a single piece of information or directly comparing two known entities without additional intermediary steps. \\
For questions that require multiple steps of reasoning or fact-finding, where each sub-question logically depends on the answer to a previous one. For these, the sub-questions should be interconnected such that once all of them are answered in sequence, the original question can be answered.\\
For questions that need to do comparison, the final sub-question should present the information from previous sub-questions using declarative sentences and reiterate the original question so that the comparison can be made based on the gathered information.

Input fields are:\\
Question: \{the input question\}\\
Output fields are:\\
Reasoning: \{reasoning for how to decompose the question\}\\
Output: \{the decomposed sub-questions\}
\end{tcolorbox}

\begin{tcolorbox}[mypromptbox, title=Prompt and Template for Question Construction]
You are given the original question, a list of answered sub-questions and their answers based on the original question.\\
For each question-answer pair, the question is labeled with a number as "Question \#n", and each corresponding answer is labeled with the same number as "Answer \#n", we also have reasonings for getting each answer, which is labeled as as "Reasoning \#n".\\
You then have a new question that may reference earlier answers using these labels "\#n".\\
Your task is to rewrite this new question by replacing every reference (like "\#1") with the corresponding answer from the list of previous questions and answers.\\
To output the good rewritten question, you should:\\
a. Identify all placeholders in the new question (such as "\#1" "\#2" etc.).\\
b. Find the corresponding answer from the list of previous question-answer pairs.\\
c. From that answer, extract the key information that logically replaces the placeholder, ignore any additional background information or explanations. This key information might be a single word or a phrase that best fits the context of the new question.\\
d. Rewrite the new question, replacing each placeholder with the key information extracted from the corresponding answer.\\
e. If the corresponding answer did not provide any useful information, please based on the original question to rewrite this sub-question, make it understandable and answerable.\\
f. If necessary, adjust the wording so the output rewritten question is grammatically correct and logically coherent.\\
g. If there are no placeholders or no relevant information to substitute, return the original new question unchanged.\\

Input fields are:\\
Original Question: \{the original question, following sub-questions are derived from this question\}\\
Question-Answer Pairs: \{the previous questions and answers\}\\
New Question: \{the new question that needs to be rewritten\}\\
Output fields are:\\
Rewritten Question: \{the rewritten output\}
\end{tcolorbox}

\begin{tcolorbox}[mypromptbox, title=Prompt and Template for Retrieval Decision]
You are given a new question. Your task is to determine whether the question already contains enough information to be answered without retrieving additional information.\\
If the question includes all the necessary details, output "false" (no retrieval needed).\\
If the question is missing information and cannot be answered as-is, output "true" (retrieval needed).\\
To output the correct decision, you should:\\
a. Check the question and analyze it thoroughly.\\
b. If everything required to answer it is explicitly stated or can be inferred from the text of the question itself, respond with "false" (meaning no retrieval needed).\\
c. Otherwise, if key information is not provided, respond with "true" (meaning retrieval is needed).\\
d. DO NOT rely on your own knowledge to answer the question. Make your decision based solely on the information provided in the question text.\\
For example:\\
Question: What is the birth date of Lily?\\
Analysis: The question does not provide a birth date, so we need additional info.\\
Output: true\\
Question: Alice was born on April 6, 2001 and Jack was born on May 15, 2002. Which person was born earlier?\\
Analysis: The question includes all needed information in the text. We just need to do a comparison to get the answer.\\
Output: false 

Input fields are:\\
Question: \{the input question\}\\
Output fields are:\\
Analysis: \{the analysis for whether needs retrieval\}\\
Output: \{true or false\}
\end{tcolorbox}

\begin{tcolorbox}[mypromptbox, title=Prompt and Template for Query Rewriting]
You are an expert in refining and optimizing retrieval queries.\\
Your task is to rewrite the current question into a concise and effective retrieval query, ensuring it captures the key intent and is optimized for precision.\\
Additionally, if "Last Rewritten Query" is provided, use it as a reference for continuity and improvement.\\
If "Last Rewritten Query" is "none", assume the original question was used as-is for retrieval and rewrite the query from scratch.\\
For example:\\
Example 1:\\
Question: What is the capital of France?\\
Last Rewritten Query: Capital city of France\\
New Query: France capital city\\
Example 2:\\
Question: What is the tallest mountain in the world?\\
Last Rewritten Query: none\\
New Query: Tallest mountain in the world\\

Input fields are:\\
Question: \{the input question\}\\
Last Rewritten Query: \{last time's rewritten query\}\\
Output fields are:\\
New Query: \{the new rewritten query\}
\end{tcolorbox}

\begin{tcolorbox}[mypromptbox, title=Prompt and Template for Passage Reranking]
You are an intelligent assistant that can rank passages based on their relevancy to the query.\\
We will provide you with a few passages, each indicated by a numerical identifier [].\\
Rank the passages in how well their content can be used to directly answer the query.\\
Only output the top-10 most relevant passages, listed in descending order of relevance.\\
If fewer than 10 passages are provided, rank all available passages.\\
Please give the reasoning for ranking the passages.\\
The final output format should be [x] > [y] > [z] > [w] > [v] and should contain exactly 10 numerical identifiers if 10 or more are available, or all available identifiers if fewer than 10 are provided.\\
You can only use ">", using "=" and "<" is prohibited. This means you should always give descending ranking, no equal and ascending ranking.\\
For example:\\
Identifiers: [1], [2], [3], [4], [5]\\
Output: [2] > [3] > [5] > [1] > [4]\\
Identifiers: [1], [2], [3], [4], [5], [6], [7], [8], [9], [10]\\
Output: [2] > [9] > [5] > [8] > [4] > [1] > [10] > [7] > [3] > [6]\\
Only respond with the ranking results, do not say any word or explain.\\

Input fields are:\\
Query: \{the input search query\}\\
Identifiers: \{all the numerical identifiers of given passages\}\\
Context: \{the passages need to be selected and ranked\}\\
Output fields are:\\
Reasoning: \{reasoning for how to select and rank the given passages\}\\
Output: \{the ranking results\}
\end{tcolorbox}

\begin{tcolorbox}[mypromptbox, title=Prompt and Template for Answer Generation]
Answer the question using only the information in the provided context.\\
You should reason through the context to get the answer. Clearly present the relevant information in your reasoning.\\
If, after careful reasoning, the context does not provide enough information to answer the question, respond with: "I don't know."\\
The question is a simplified step (one hop) derived from a more complex question.\\
To help remove ambiguity, we may provide background information including the original full question and previous answers to previous related sub-questions.\\
Use this background only to understand the intent of the question—do not extract facts or cite anything from it. Background may be empty in some cases.\\
You MUST include a citation for **every factual statement** in your answer using the context number(s) in square brackets immediately after the relevant information.\\
Only omit citations if the context does not support any part of the answer and you respond with: "I don't know."\\
Citation format: [number], or [number1][number2] if supported by multiple contexts.\\
Examples:\\
Answer: He went to school at 8 am [1].\\
Answer: She wrote book A [2] and book B [4].\\
Answer: They are shopping for food [1][3].\\
Your answer must be accurate, direct, complete, and concise.\\
Do not include additional background or explanation beyond what is necessary to answer the question.\\

Input fields are:\\
Question: \{the question needs to be answered\}\\
Context: \{may contain relevant information\}\\
Background: \{the original question and other question hops with answers\}\\
Output fields are:\\
Reasoning: \{reasoning to output the answer\}\\
Output: \{the output answer\}
\end{tcolorbox}

\begin{tcolorbox}[mypromptbox, title=Prompt and Template for Answer Verification]
You are a verifier. Your task is to determine whether the answer correctly solves the question and is well-supported by the cited context(s).\\
You are given a question, an answer, and a set of citations. Follow this two-step process:\\
Step 1: Answer Quality\\
Determine whether the answer correctly, directly, and fully answers the question.\\
It must not be vague, evasive, or incomplete.\\
If the answer fails to provide a valid, specific response to the question: for example, by saying "not enough information" or "I don't know", you must output "false".\\
Step 2: Citation Support (only if Step 1 passes)\\
Check that if every claim in the answer is explicitly supported by the cited context(s).\\
The answer must not include information that is missing or unsupported by the citations.\\
Always provide a brief explanation in the reason section.\\
Please output your final judgment as either "true" (the answer is correct and fully supported) or "false" (the answer is incorrect or not supported).\\

Input fields are:\\
Question: \{the input question\}\\
Answer: \{the given answer\}\\
Source: \{the associate citations\}\\
Output fields are:\\
Reason: \{reason of choosing true or false\}\\
Output: \{true or false\}
\end{tcolorbox}

\begin{tcolorbox}[mypromptbox, title=Prompt and Template for Final Answering]
A multi-hop reasoning question is one where the answer cannot be found by looking at a single piece of information or taking just one step.\\
Instead, it requires combining information from multiple sources or completing several logical steps.\\
To tackle such questions, we need to decompose the main question into smaller, more manageable sub-questions.\\
By answering these sub-questions one by one, you build up to the final answer.\\
Now we will provide a multi-hop reasoning question, with a set of its decomposed sub-questions.\\
We also provide the reasonings and answers to each sub-questions.\\
Your task is to combine these pieces of information to answer the original main question thoroughly and accurately.\\
Please articulate your reasoning for the final answer.\\
To achieve the good output, you should:\\
a. Read the sub-questions with their reasonings and answers carefully.\\
b. Integrate or synthesize the answers to the sub-questions into a single, coherent answer to the original question.\\
c. Ensure your final answer is simple, clear, concise, and correct.\\

Input fields are:\\
Question: \{the original multi-hop reasoning question\}\\
Decomposed Information: \{the answered sub-questions\}\\
Output fields are:\\
Reasoning: \{reasoning to output the answer\}\\
Answer: \{the output answer\}
\end{tcolorbox}

\begin{tcolorbox}[mypromptbox, title=Prompt and Template for Improve Analysis in Self-Reflection]
You are an expert in multi-hop question answering and decomposition.\\
Your task is to analyze a provided question decomposition and offer high-level guidance on how to correct and improve it, if necessary.\\
Use the rewritten sub-questions, reasoning, and answers only to detect decomposition flaws—not to inform your revised logic.\\
You will receive:\\
a. A complex question requiring multi-hop reasoning.\\
b. A record of sub-questions, including: the original decomposition, rewritten sub-questions, reasoning steps, and answers.\\
Assume the current decomposition may be incorrect. Provide structured, bullet-pointed instructions on how to improve it by:\\
a. Proposing a logical, step-by-step breakdown of the question.\\
b. Ensuring sub-questions extract critical information in the right order.\\
c. Avoiding dependency errors or propagation of uncertain information.\\
If the decomposition is already sound and each sub-question is clearly answerable, respond with:\\
The previous decomposition is appropriate. Please follow the same structure.\\
Do not provide new sub-questions. Please only give high-level instructions on how the question should be decomposed.\\

Input fields are:\\
Question: \{the input question\}\\
Sub-questions Solving Record: \{the previous decomposed sub-questions and their solving process\}\\
Output fields are:\\
Analysis: \{instructions for the new decomposition logic\}
\end{tcolorbox}

\begin{tcolorbox}[mypromptbox, title=Prompt and Template for Improve Decomposition in Self-Reflection]
You are an expert in multi-hop reasoning and question decomposition.\\
Your task is to analyze the original question and the previous one or multiple decompositions, which have been determined to be incorrect, and generate a new, logically sound decomposition that leads to the correct final answer.\\
Important instructions:\\
a. The previous decomposition(s) are incorrect and led to wrong final answers.\\
b. Do not rely on the structure of the previous decomposition(s) when generating a new one. Instead, construct a fresh, logically valid decomposition directly from the original question.\\
c. If multiple incorrect decompositions are provided, analyze and identify the reasoning flaws in each before generating the new decomposition.\\
d. You can use the instructions for help when constructing the new decomposition.\\
e. Provide reasoning that explains the logic behind the new decomposition. The new decomposition logic should be different from the previous ones.\\
f. Ensure that the new sub-questions follow a proper reasoning order and can be answered step-by-step to reach the final answer.\\
g. If a sub-question depends on a previous answer, explicitly reference it using "\#number" rather than vague phrasing like "based on".\\
For example:\\
Question: Which city is home to the headquarters of the company founded by the inventor of the telephone?\\
Previous decomposition:\\
Incorrect decomposition 1:\\
1. Who invented the telephone?\\
2. Which company was founded by \#1?\\
3. Which city is home to the headquarters of \#2?\\
New decomposition instructions 1:\\
The second step is too vague. Instead of asking about where the company is located, specify that we need the **headquarters** location.\\
Reasoning: The previous decomposition jumps directly from identifying the company to identifying the city, skipping a crucial intermediate step. Before determining which city the headquarters is in, we should first establish the exact headquarters location of the company. This makes the reasoning more structured and avoids potential errors if a company has multiple locations.\\
New decomposition:\\
1. Who invented the telephone?\\
2. Which company was founded by \#1?\\
3. Where are the headquarters of \#2?\\
4. Which city is home to \#3?\\

Input fields are:\\
Question: \{the input question\}\\
Previous Decompositions: \{the previous decomposed sub-questions and the instructions for improving\}\\
Output fields are:\\
Reasoning: \{reasoning to the new decomposition\}\\
New Decomposition: \{the new decomposition based on the question\}
\end{tcolorbox}

\begin{tcolorbox}[mypromptbox, title=Prompt and Template for Final Answer Verification]
You are a verifier tasked with determining whether a given answer is correct and well-supported by the provided passages.\\
Given a question, an answer, and the supporting passages, evaluate the answer based on the following criteria:\\
a. The answer must be factually correct based on the provided passages.\\
b. The answer must fully and directly address the question without omitting critical details necessary for a comprehensive response.\\
c. The answer must be explicitly substantiated by the provided passages, meaning that every claim made in the answer must be directly supported by the passages without requiring external knowledge.\\
d. If the passages do not contain enough information to support the answer, or if the answer includes information not found in the passages, the answer should be considered incorrect.\\
e. The question always has an answer. If no valid answer can be determined from the passages, it means either the question itself has problems or the necessary supporting passages are missing. In this case, the answer is incorrect by default. You MUST output false, no exceptions.\\
You need to provide a brief justification explaining why the answer is correct or incorrect based on the passages in the reason section.\\
Output true if the answer is factually accurate, relevant, complete, and fully supported by the passages. Otherwise, output false.\\

Input fields are:\\
Question: \{the input question\}\\
Answer: \{the given answer\}\\
Passages: \{the supporting passages for the answer\}\\
Output fields are:\\
Reason: \{why the answer is correct or incorrect\}\\
Output: \{true or false\}
\end{tcolorbox}

\begin{tcolorbox}[mypromptbox, title=Prompt and Template for LLM Evaluation]
You are an evaluator responsible for determining whether a given answer correctly aligns in meaning with a provided ground truth answer.\\
Evaluation Criteria:\\
a. Relevance: The answer must directly address the given question and provide a coherent, contextually appropriate response.\\
b. Semantic Alignment: Compare the answer with the ground truth(s) to ensure they convey the same core meaning or intent, even if phrased differently.\\
c. Completeness: The answer must include all essential information present in the ground truth.\\
Ground Truth Format:\\
a. The ground truth is provided as a list, containing one or multiple valid expressions of the correct answer.\\
b. If multiple ground truths are given, they are semantically equivalent but differ in wording.\\
Output Instructions:\\
a. Output true if the answer aligns in meaning with at least one of the ground truth expressions and includes all key details.\\
b. Output false if the answer fails to align with any of the ground truths or omits important information.\\

Input fields are:\\
Question: \{the input question\}\\
Ground Truth Answer: \{the ground truth answer\}\\
Our Answer: \{the answer needs to be evaluated\}\\
Output fields are:\\
Reasoning: \{reasoning to determine whether our answer is correct compared to the ground truth\}\\
Output: \{true or false\}
\end{tcolorbox}

\end{document}